\documentclass[10pt,logo]{nvidiatechreport}
\usepackage[authoryear,sort&compress,round]{natbib}
\usepackage{parskip}
\usepackage{amsmath,amssymb,amsfonts,bm}
\usepackage{graphicx}
\usepackage{booktabs}
\usepackage{multirow}
\usepackage{makecell}
\usepackage{array}
\usepackage{tabularx}
\usepackage{colortbl}
\usepackage{xcolor}
\usepackage{subcaption}
\usepackage{wrapfig}
\usepackage{placeins}
\usepackage{microtype}
\usepackage{url}
\usepackage{xspace}
\usepackage{tikz}
\usepackage{pgfplots}
\pgfplotsset{compat=1.18}
\usetikzlibrary{arrows.meta,positioning,fit,backgrounds,calc,shapes.geometric,shapes.misc,decorations.pathreplacing,patterns}
\usepackage[capitalize]{cleveref}
\usepackage{enumitem}
\usepackage{soul}

\newcommand{\method}{\mbox{LocateAnyMesh}\xspace}
\newcommand{\methodshort}{\mbox{LAM}\xspace}
\newcommand{\lamdirect}{\mbox{\methodshort\unskip-Direct}\xspace}
\newcommand{\lamselect}{\mbox{\methodshort\unskip-Select}\xspace}

\definecolor{cbBlue}{HTML}{0072B2}
\definecolor{cbOrange}{HTML}{E69F00}
\definecolor{cbGreen}{HTML}{009E73}
\definecolor{cbRed}{HTML}{D55E00}
\definecolor{cbPurple}{HTML}{CC79A7}
\definecolor{cbSky}{HTML}{56B4E9}
\definecolor{cbYellow}{HTML}{F0E442}
\definecolor{cbGray}{HTML}{666666}
\definecolor{panelBG}{HTML}{F4F6F8}
\definecolor{goodGreen}{HTML}{0B7A45}
\definecolor{badRed}{HTML}{B02418}

\newcommand{\best}[1]{\textbf{#1}}

\newcommand{\na}{\textendash}
\newcommand{\notmeasured}{\textit{n/e}}

\newcommand{\cmark}{\textcolor{goodGreen}{\ding{51}}}

\usepackage{pifont}

\renewcommand{\titlefont}{\color{nvidiagreen}\normalfont\bfseries\fontsize{15}{18}\selectfont}
\renewcommand{\absfont}{\bfseries\linespread{1.05}\fontsize{10}{12}\selectfont}
\titlespacing*{\section}{0pt}{2ex plus 2pt minus 1pt}{4pt}
\titlespacing*{\subsection}{0pt}{1.5ex plus 2pt minus 1pt}{2pt}
\titleformat{\paragraph}[runin]{\normalfont\normalsize\bfseries}{}{0pt}{#1}
\titlespacing*{\paragraph}{0pt}{0.6ex plus 0.3ex minus .2ex}{0.5em}

\newcommand{\tokens}{\mathbf{z}}
\newcommand{\discrim}{\textsc{Discrim}\xspace}

\title{\method: Grounding 3D Mesh Queries in 2D Images with Vision-Language Models}

\author{Chris Choy, Zhiding Yu, Jan Kautz}
\affil{NVIDIA Research}
\hypersetup{pdfauthor={Chris Choy, Zhiding Yu, Jan Kautz}}

\begin{document}
\maketitle

\noindent\begin{minipage}{\textwidth}
\centering
\includegraphics[width=\linewidth]{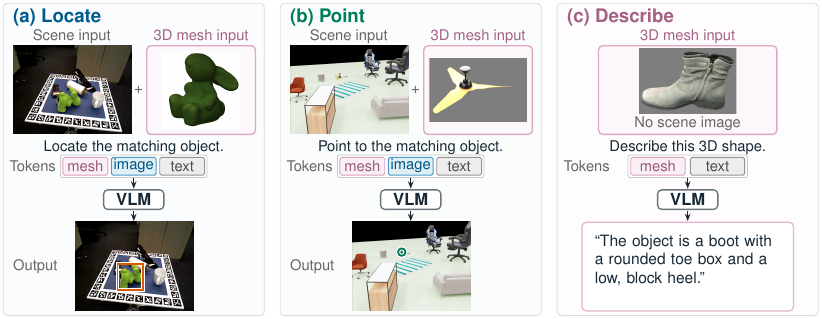}
\captionsetup{font=normalsize,skip=6pt,hypcap=false}
\captionof{figure}{\textbf{A 3D mesh is a direct input to a VLM.}
Mesh tokens encode geometry and surface color; mesh views visualize the assets for the reader.
\textbf{(a)} Localization in an unseen HomebrewedDB scene \citep{kaskman2019hb}.
\textbf{(b)} Pointing to the queried fan in a synthetic ABO scene.
\textbf{(c)} A literal first-sentence excerpt of a mesh-only description.
All three panels use \lamdirect, initialized from LocateAnything-3B:
a localization checkpoint for (a) and a separate multitask checkpoint for (b,c).
These selected illustrations are not performance estimates; errors and controlled
evaluations are reported in \cref{sec:capabilities}.}
\label{fig:hero}
\end{minipage}\par

\begin{abstract}
Textured 3D meshes specify object geometry and appearance, making them a natural query for
finding a particular asset in an image. We introduce \method{}, a vision--language model
that uses the mesh itself as part of its prompt. A frozen 3D encoder maps colored surface
points to a compact token sequence, which a learned connector places alongside image and
instruction tokens. We evaluate two grounding models with PPAT mesh tokens:
\lamselect selects image-region proposals, while
\lamdirect predicts coordinates
directly. Neither requires rendered query views or view-template matching. New objects
require a single encoder pass and no object-specific fine-tuning. \lamselect improves localization
on the reported real-image suites and the official BOP YCB-V evaluation.
Secondary experiments compare mesh encoders, and a \lamdirect study isolates
surface color. A single-seed \lamselect continuation supplies pointing and
mesh-only description evidence without replacing the reported localization model.
The results support mesh tokens as a query representation, with remaining limitations in
presence decisions, identity binding, proposal coverage, and descriptive fidelity.
\end{abstract}

\abscontent
\section{Introduction}
\label{sec:intro}

Grounding VLMs such as Florence-2, Rex-Omni, and LocateAnything generate spatial coordinates
from images and language prompts \citep{xiao2024florence2,jiang2025rexomni,wang2026locateanything}.
A textured mesh specifies an asset's geometry and surface color directly. A phrase such as
\emph{the green figurine} may identify one target in \cref{fig:hero}, but distinguishing similar
assets in words can require a much longer query. We ask whether a vision--language model can use
the mesh itself to locate an asset in a scene or describe it.

\begin{figure}[t]
\centering
\includegraphics[width=\textwidth]{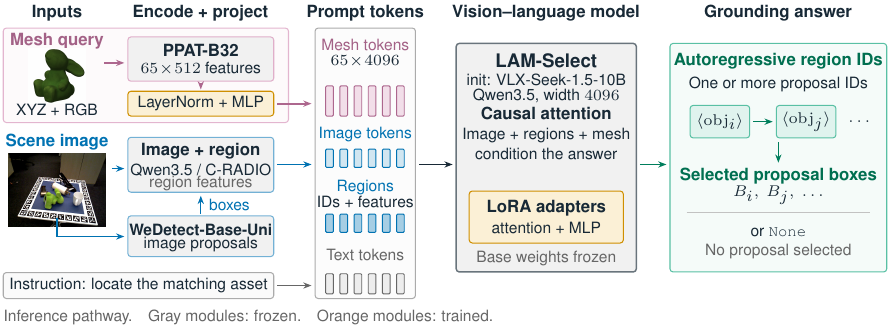}
\caption{\textbf{\lamselect, initialized from VLX-Seek-1.5-10B.} A frozen 3D encoder and a
trainable connector place mesh features alongside image, region, and instruction tokens.
An image-only detector supplies candidate boxes. The Qwen3.5 decoder generates region IDs or an explicit
refusal; selected IDs map back to the candidate boxes. Gray modules are frozen;
orange modules are trained.}
\label{fig:pipeline}
\end{figure}

\method{} conditions a grounding VLM on direct mesh input. We compare two variants,
\lamdirect and \lamselect, initialized from the released LocateAnything-3B
\citep{wang2026locateanything} and VLX-Seek-1.5-10B \citep{omailab2026vlxseek}
grounding checkpoints, respectively. Both use PPAT mesh tokens.
A frozen OpenShape-RGB PointBERT encoder
\citep{liu2023openshape,yu2022pointbert} maps the colored surface to $65$ tokens; a trainable
MLP projects them into the same sequence as the instruction and any image tokens
(\cref{fig:pipeline}). We name the variants by their output mechanism:
\lamdirect generates box coordinates, whereas \lamselect chooses among detector proposals.

\begin{table}[!htbp]
\centering
\footnotesize
\setlength{\tabcolsep}{3pt}
\begin{tabularx}{\textwidth}{@{}l>{\raggedright\arraybackslash}X>{\raggedright\arraybackslash}Xc>{\raggedright\arraybackslash}X>{\raggedright\arraybackslash}X>{\raggedright\arraybackslash}X@{}}
\toprule
Variant & Base model & LLM & Params. & Vision encoder(s) & Output & External detector at eval. \\
\midrule
\lamdirect & LocateAnything-3B & Qwen2.5-3B-Instruct & 3B class & MoonViT-SO-400M & Box coordinates & None \\
\lamselect & VLX-Seek-1.5-10B & Qwen3.5 & 10B class & Qwen3.5-VL + C-RADIO & Region references & WeDetect-Base-Uni, top 100 \\
\bottomrule
\end{tabularx}
\caption{\textbf{The two \method{} variants, named by output mechanism.}
Parameters are released full-system size classes, not adapter-inclusive audits.
Both variants use frozen PPAT mesh tokens.}
\label{tab:variants}
\end{table}

Model-based detectors also accept a CAD asset and an image, but render the asset into views and
match their descriptors to image regions \citep{nguyen2023cnos,lin2024sam6d,tang2026cadsam3}.
We feed mesh tokens directly to the language decoder. \lamselect selects proposals from an image-only
detector; the query mesh does not need rendered views. Each new object requires one encoder pass.

Our main localization benchmark compares both systems on held-out HOPE, LM-O, and HB
objects and on seen YCB-V objects. Every YCB-V detection AP uses the official $900$-image
BOP target list with amodal scoring (\cref{sec:results}).

\begin{figure}[p]
\centering
\includegraphics[width=0.85\textwidth]{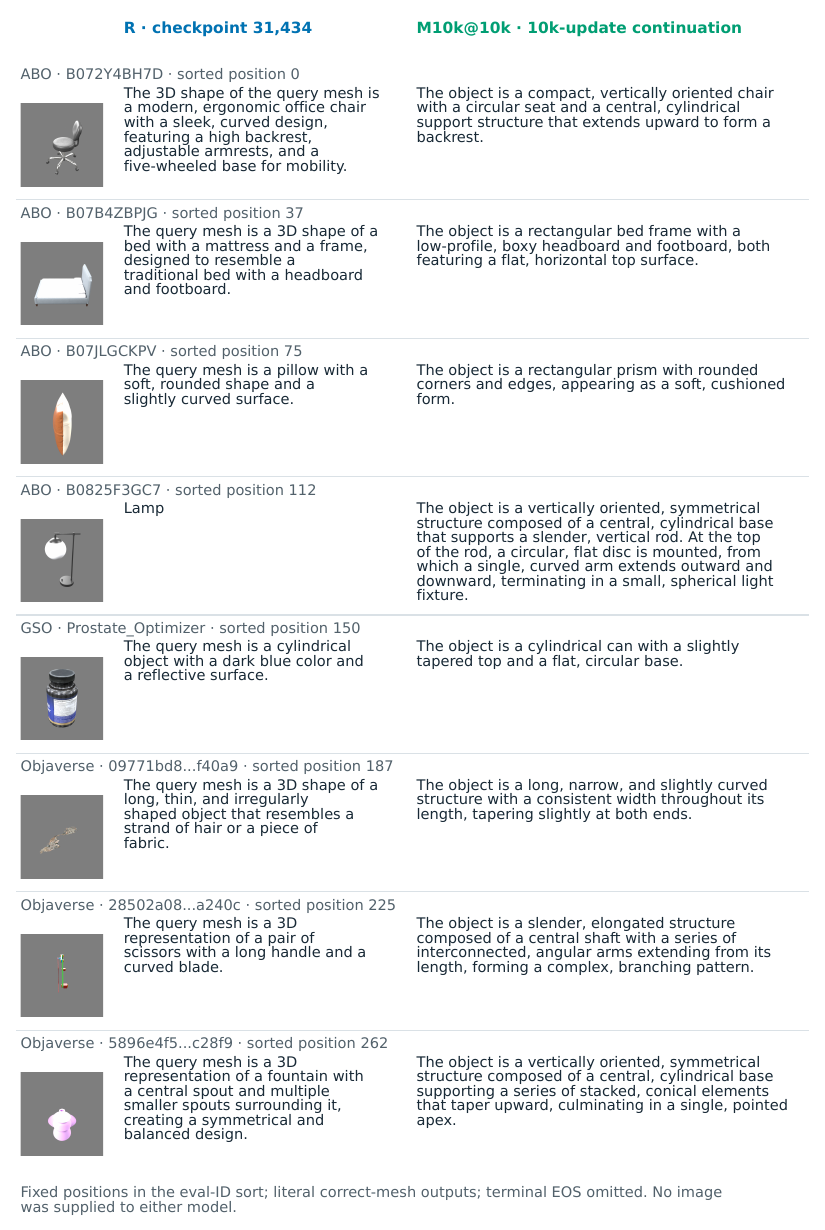}
\caption{\textbf{Mesh-only descriptions from R and M10k@10k.}
Fixed before output inspection: zero-based positions $0,37,75,112,150,187,225,262$
in the evaluation-ID sort of all $300$ identities, without quality ranking.
Each row shows \texttt{view\_0} for the reader and the literal correct-mesh responses;
only terminal EOS is omitted. The models receive mesh tokens and the instruction,
but no image or render.}
\label{fig:vlx-m10k-descriptions}
\end{figure}

\begin{figure}[p]
\centering
\includegraphics[width=0.85\textwidth]{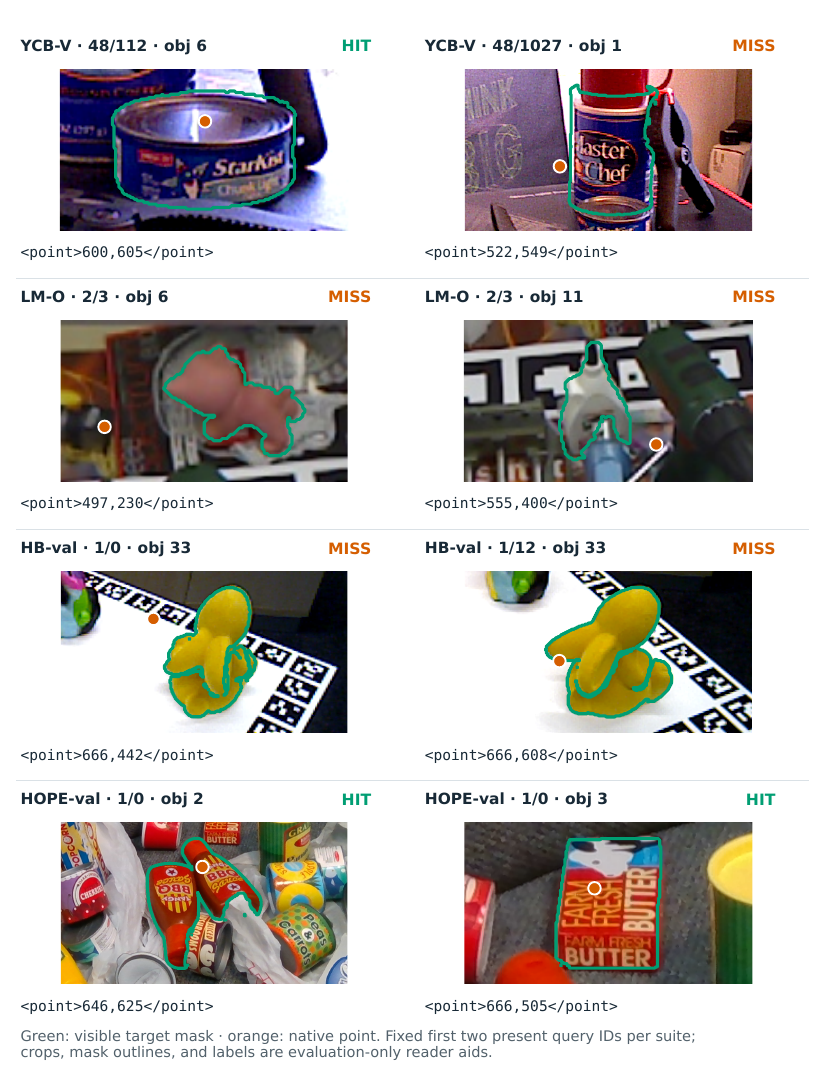}
\caption{\textbf{M10k@10k native pointing on real BOP images.}
The fixed gallery takes the first two present query IDs in each suite after
sorting, without using outcomes: three are hits and five are misses. Text is the
literal model response with only terminal EOS omitted. Coordinates use the
size-minus-one pixel convention. Orange points and green visible-mask outlines
determine the displayed hit/miss labels; the crops, masks, and labels are
evaluation-only reader aids and were not model inputs.}
\label{fig:vlx-m10k-pointing}
\end{figure}

A single-seed multitask continuation of the \lamselect localization checkpoint
provides the main pointing and mesh-only description studies
(\cref{fig:vlx-m10k-descriptions,fig:vlx-m10k-pointing}). It is used only for
capability evidence; the localization tables retain the original checkpoint.
Separate \lamdirect capability studies and the original \lamselect transfer
probes remain in \cref{sec:capabilities}.

\paragraph{Contributions.}
\begin{itemize}[leftmargin=1.4em,itemsep=1pt,topsep=2pt]
\item \textbf{Direct mesh input to two grounding VLMs.} A shared PPAT query representation
conditions proposal selection or coordinate generation through a learned connector
(\cref{sec:method}).
\item \textbf{Representation comparisons and grounding errors.} Within \lamselect, we compare
PPAT and Uni3D-S and audit proposal, presence, and identity errors. A matched \lamdirect
intervention tests query color (\cref{sec:vlxencoders,sec:analysis,sec:iso1}).
\end{itemize}

\section{Related Work}
\label{sec:related}

\paragraph{Generative grounding VLMs.}
Florence-2 \citep{xiao2024florence2} unifies captioning, detection, and grounding through task
prompts and sequence generation. Rex-Omni \citep{jiang2025rexomni} combines coordinate-token
outputs with text and image-region prompts. LocateAnything \citep{wang2026locateanything}
predicts the coordinates within each box in parallel. VLX-Seek \citep{omailab2026vlxseek}
uses region proposals for grounding. \method{} extends this line of work on the input side:
native mesh tokens condition proposal selection in \lamselect and direct-coordinate
generation in \lamdirect.

\begin{figure}[tb]
\centering
\includegraphics[width=\textwidth]{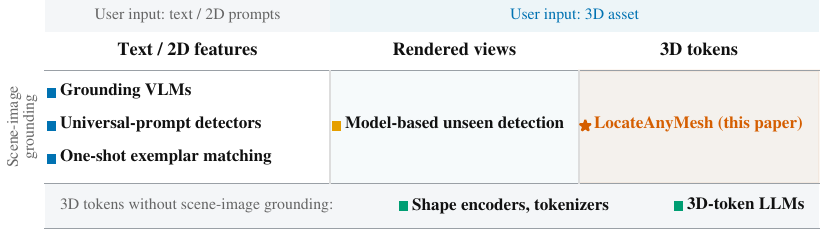}
\caption{\textbf{Same CAD input, different network representations.} Model-based detection and
\method{} ground a supplied CAD asset in a scene image using rendered views and native 3D tokens,
respectively.}
\label{fig:queryfamilies}
\end{figure}

\Cref{fig:queryfamilies} separates the supplied query from its network representation.
CNOS \citep{nguyen2023cnos}, SAM-6D \citep{lin2024sam6d}, and CAD-Prompted SAM3
\citep{tang2026cadsam3} accept a CAD asset and a scene image, a setting evaluated by BOP
\citep{hodan2024bop2023}, but use rendered query views. CAD-Prompted SAM3 returns
instance masks that also define boxes. Universal-prompt detectors
\citep{wu2024glee,carion2025sam3} accept 2D prompts. Exemplar matching methods such as
OS2D and the visual-prompt modes of T-Rex2 condition on 2D examples
\citep{osokin2020os2d,jiang2024trex2}.
PointLLM \citep{xu2024pointllm} generates language from colored point clouds, while
Kyvo \citep{sahoo2025kyvo} models structured 3D scenes and evaluates both synthetic
and real-world settings. Here, the supplied 3D asset serves as a query for image grounding.

\section{Method}
\label{sec:method}

\paragraph{Task.}
\label{sec:task}

Mesh-prompted grounding takes an RGB image and a textured 3D asset as input, and
localizes matching instances with boxes or explicitly refuses an absent query.
Queries identify individual assets, including objects unseen during grounding training.

We implement \method{} with two pretrained grounding systems, both using frozen PPAT mesh tokens.
\lamdirect initializes from the released LocateAnything-3B checkpoint
\citep{wang2026locateanything}, inheriting its Qwen2.5-3B-Instruct decoder, MoonViT image
encoder, and direct-coordinate output. \lamselect initializes from the released
VLX-Seek-1.5-10B checkpoint \citep{omailab2026vlxseek}, inheriting its Qwen3.5 text decoder,
image towers, region encoder, and autoregressive region-ID output.
A learned connector adds mesh features to each system's input sequence.
We name the variants by output mechanism: \lamdirect generates box coordinates directly,
whereas \lamselect selects detector proposals. They also differ in base model, scale,
language model, vision towers, and training; this is a complete-system comparison,
not an isolated language-model ablation.
\Cref{tab:tokens,tab:hparams} compare their output representations and training settings.

\paragraph{From a textured mesh to 65 tokens.}
We sample ten thousand colored points from the mesh surface and normalize their coordinates to
the unit sphere. A frozen OpenShape-RGB PointBERT encoder, PPAT-B32
\citep{liu2023openshape,yu2022pointbert}, maps their coordinates and RGB values to $65$ tokens.
A trainable LayerNorm--MLP connector maps each width-$512$ token to the \lamselect decoder's
width of $4{,}096$. A secondary Uni3D-S \citep{zhou2024uni3d} comparison is reported
in \cref{sec:vlxencoders}. New assets require one encoder
pass; its tokens can be cached for reuse. Mesh onboarding is excluded from downstream
inference timings and remains unmeasured on GB200 (\cref{sec:onboardingbench}).

\paragraph{Image regions and decoding.}
At inference, WeDetect-Base-Uni supplies up to $100$ image-only candidate boxes. \lamselect's
frozen image towers and native region encoder produce a feature token for each candidate.
The language decoder receives image context, region IDs and features, the mesh span, and
the instruction. Causal autoregression emits one or more region IDs, or \texttt{None}.
Each selected ID returns its proposal's coordinates. The mesh is never rendered for the
model, and ground-truth boxes are not used to form inference candidates. Localization
recall is limited by proposal coverage as well as the decoder's selections.

\paragraph{\lamselect training.}
We train the mesh connector and language-model LoRA adapters \citep{hu2022lora}; the base
weights, image towers, region encoder, and mesh encoder remain frozen. Training prompts
request all matching instances, with explicit absent queries. Training candidates are
ground-truth visible boxes, and the objective is causal cross entropy on the answer and
termination tokens. This creates a candidate-distribution difference from evaluation,
which uses detector proposals. The \lamselect reference uses $31{,}434$ updates;
\cref{sec:vlximplementation} gives its training settings and encoder control.

\section{Experimental Setup}
\label{sec:data}

\subsection{Training data}

\begin{figure}[t]
\centering
\includegraphics[width=\textwidth]{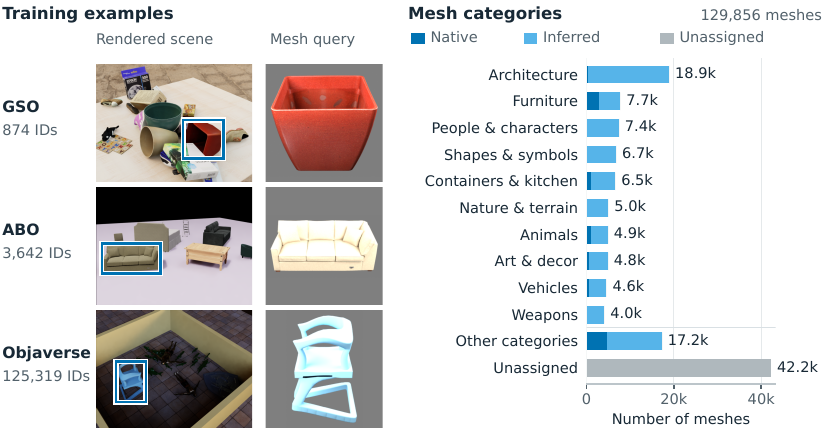}
\caption{\textbf{Training examples and mesh categories.} Examples and the asset inventory
come from the earlier \lamdirect data analysis. Blue boxes mark training targets;
GSO meshes are scanned. The histogram counts each eligible mesh ID once across all sources,
including YCB-V. Native labels are supplemented by categories inferred from descriptions;
smaller groups form ``Other categories'' (\cref{sec:datacomposition}).}
\label{fig:training-data}
\end{figure}

We train on Google Scanned Objects \citep{downs2022gso}, Amazon Berkeley Objects
\citep{collins2022abo}, two UID-disjoint Objaverse~1.0 tabletop waves
\citep{deitke2023objaverse}, and BOP YCB-V
\citep{xiang2018posecnn,hodan2018bop,denninger2023blenderproc2}.
The \lamselect recipe also includes an Objaverse PPAT union. Its PPAT and Uni3D
runs share query order and an absent-sampling probability of $0.25$, with localization
supervision only. Source mixtures and sampling details are in
\cref{tab:vlxdata,sec:datacomposition}; \cref{fig:training-data} shows examples and mesh categories.

\paragraph{Held-out splits and contamination.}
Synthetic recipes define object-level held-out splits; YCB-V retains its $21$ seen identities.
An embedding-space audit finds duplicate assets across training and evaluation IDs in all
three simulated suites and may miss others (\cref{sec:datacomposition}).
The retained synthetic scores use the original splits, so they do not establish
fully held-out identity transfer. The audit flags no pairs in the real BOP suites;
BOP suites other than YCB-V are evaluation-only.

\subsection{Evaluation protocol}
\label{sec:protocol}

We evaluate HOPE \citep{tyree2022hope}, LM-O \citep{brachmann2014lmo},
HB \citep{kaskman2019hb}, and YCB-V. \Cref{tab:suites,sec:samplingdetail} give
the cohorts and sampling rules. Each query pairs an image with a mesh; we query
present classes and up to $8$ seeded absent classes per image, or all $21$ models on YCB-V.

\paragraph{Detection and scoring.}
\lamselect uses greedy autoregressive proposal selection and retains all selected boxes;
a top-one diagnostic keeps the highest-scored selection. \lamdirect uses the
single-instance prompt and caps decoding at one box. Invalid outputs contribute no
detections but count as failures, not valid refusals. The reported \lamselect
localization evaluations contain no invalid answers.

We report COCO AP@[.50:.05:.95], macro-averaged over classes \citep{lin2014coco}.
AP confidence comes from decoder-token probabilities, not detector scores, and is not
calibrated as a presence probability (\cref{sec:implementation}).
The house protocol uses modal boxes for instances at least $10\%$ visible, pooling
scores per dataset. Official YCB-V results use the $900$-image target list, mask-derived
amodal boxes, and ignored low-visibility instances (\cref{sec:bopprotocol}).
Only this official protocol supports comparison with BOP leaderboard AP;
house scores use different images and box targets.

We also report present-target hits and absent-query false positives. The supplemental
presence-discrimination ratio excludes HB development scene \texttt{000001}
(\cref{eq:discrim}).

\paragraph{Uncertainty.}
Neither headline system has a multi-seed evaluation. Results are tied to their checkpoint
and inference route, without seed-based confidence intervals; small rerun differences
do not estimate training uncertainty.

\section{Results}
\label{sec:results}

\subsection{Localization and inference cost}
With frozen PPAT mesh tokens, \lamselect improves house mAP over \lamdirect
on the three held-out real-image suites (\cref{tab:vlxmain}). This comparison
changes backbone, proposals, training exposure, and output cardinality; it
evaluates complete systems rather than isolating the backbone. The secondary
PPAT--Uni3D-S comparison is in \cref{sec:vlxencoders}.

\begin{table}[!htbp]
\centering\small\setlength{\tabcolsep}{5pt}
\caption{\textbf{Two grounding systems with PPAT mesh tokens.}
House visible-box COCO mAP on $50$ HOPE
and $300$ images per other suite; HB includes all $13$ scenes.
\lamselect uses optimized autoregressive inference and retains all selected boxes;
its top-one HOPE mAP is $0.319$. Other \lamselect cells are unchanged
by top-one filtering on this cohort. \lamdirect uses one-box decoding; image/query
inputs and the scorer are matched on all three suites. These are internal-protocol results;
\cref{tab:vlxycbvfull} reports official BOP YCB-V separately.}
\label{tab:vlxmain}
\label{tab:vlxvariant}
\begin{tabular}{@{}lrrr@{}}
\toprule
System & HOPE & LM-O & HB \\
\midrule
\lamdirect & 0.155 & 0.181 & 0.303 \\
\lamselect & \best{0.371} & \best{0.388} & \best{0.366} \\
\bottomrule
\end{tabular}
\end{table}

\paragraph{Official BOP YCB-V.}
On the official $900$-image target list, \lamselect reaches amodal AP of $0.7152$ with PPAT,
compared with $0.6384$ for \lamdirect with the same mesh encoder
(\cref{tab:vlxycbvfull}). YCB-V identities are seen during training, so these
results measure test-image performance rather than transfer to unseen objects.

\begin{table}[!htbp]
\centering\small\setlength{\tabcolsep}{4pt}
\caption{\textbf{Official BOP YCB-V detection with PPAT.} Both systems use the same $900$ target images
across $12$ scenes and query all $21$ objects per image ($18{,}900$ queries).
Ground truth uses mask-derived amodal boxes and BOP low-visibility ignore handling.
AP is COCO AP@[.50:.05:.95]. \lamselect uses its $31{,}434$-update checkpoint; \lamdirect
uses arm~A at $45{,}823$ updates. These complete-system comparisons are not compute-matched.
Timing is the reported GB200 batch-$21$ pilot on eight images: approximate
seconds/image is $21$/median queries/s, the median of three repeat-level cohort
means; the range spans those repeats. Timing and AP use different cohorts;
\lamdirect timing uses $10{,}000$ updates, and \lamselect timing uses cached
proposals. The full encoder comparison is in \cref{tab:vlxencoder-bop}; timer
boundaries and malformed-answer accounting are in \cref{sec:blackwell-timing}.}
\label{tab:vlxycbvfull}
\label{tab:ycbv-bop}
\begin{tabular}{@{}lrrrrr@{}}
\toprule
Grounding system & AP & AP50 & AP75 & \makecell{GB200 s/img\\$b=21$, $\downarrow$}
& \makecell{Repeat range\\s/img} \\
\midrule
\lamdirect & 0.6384 & 0.7885 & 0.7112 & \best{0.322} & 0.309--0.324 \\
\lamselect & \best{0.7152} & \best{0.9150} & \best{0.8183} & 0.682 & 0.652--0.837 \\
\bottomrule
\end{tabular}
\end{table}

\begin{figure}[t]
\centering
\includegraphics[width=\textwidth]{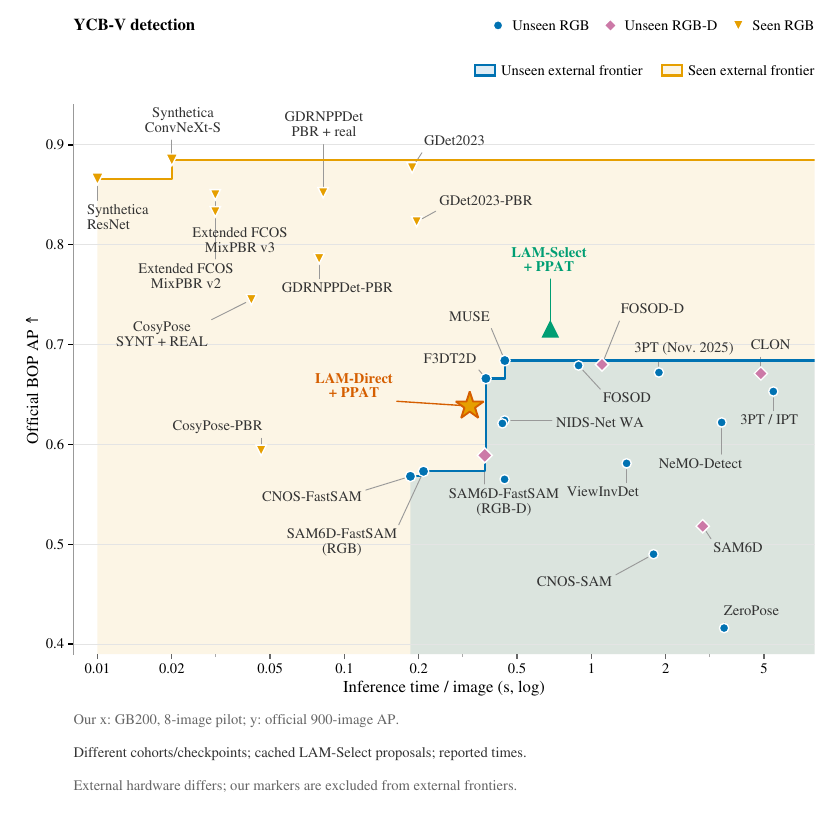}
\caption{\textbf{YCB-V accuracy and runtime context with Blackwell timings for our models.}
Seen- and unseen-object leaderboard points have separate external-only Pareto
frontiers. Our PPAT models use official
$900$-image BOP AP on the vertical axis and reported GB200 batch-$21$ time from
the eight-image pilot on the horizontal axis ($21$/median queries/s).
These coordinates come from different cohorts; \lamdirect also uses different
checkpoints for accuracy ($45{,}823$ updates) and timing ($10{,}000$).
\lamselect uses $31{,}434$ updates for both, with cached proposals in the timed pilot.
External methods retain their reported hardware and timer boundaries.
Our models saw YCB-V objects during training and are excluded from both frontiers.
This is reference context, not a controlled speed--accuracy comparison.
\Cref{tab:vlxencoder-bop} reports Uni3D-S accuracy and timing.}
\label{fig:pareto}
\end{figure}

\paragraph{Blackwell inference cost.}
On the reported eight-image GB200 pilot, batching increases throughput for both
systems (\cref{tab:blackwell-timing,tab:blackwell-batches}). Timings exclude mesh
encoding and live proposal detection; \lamselect uses cached proposals.
\Cref{fig:pareto} combines these pilot times with the separate official AP
evaluation. For \lamdirect, accuracy and timing also use different checkpoints.

Several detectors on BOP's seen-object board \citep{hodan2024bop2023} report higher
AP and lower runtime than our models, with different hardware and timer boundaries.
Our models saw YCB-V objects during training; their AP does not establish
unseen-object leaderboard superiority. Transfer evidence comes from the held-out
real suites (\cref{tab:vlxmain}); leaderboard scope is in \cref{sec:readingfig2}.

\begin{figure}[!htbp]
\centering
\includegraphics[width=0.9\textwidth]{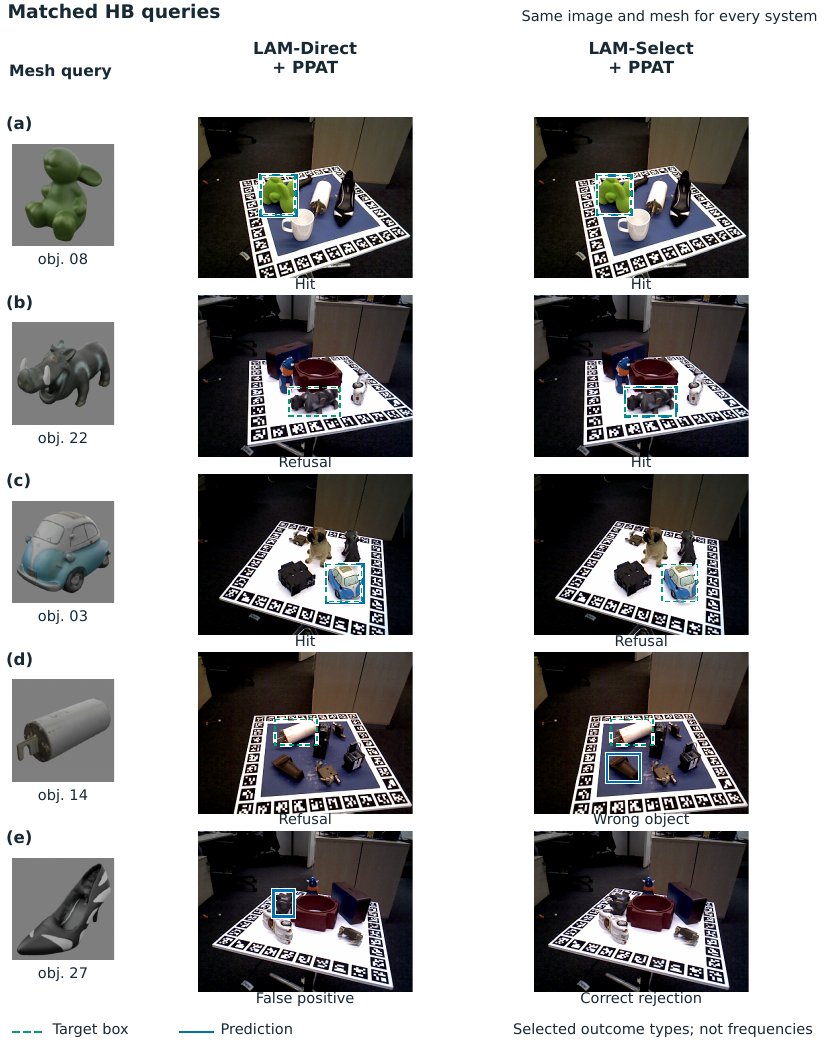}
\caption{\textbf{Two grounding systems on matched held-out HB queries.}
\lamdirect and \lamselect both use PPAT.
Every column uses the same
RGB image and mesh query. Blue boxes are predictions; dashed green boxes are visible targets.
All predicted boxes are shown, and hits require IoU $\ge0.5$. Rows illustrate shared success,
\lamselect-only success, \lamdirect-only success, a \lamselect wrong-object selection, and absent-query
disagreement. They were chosen by a fixed median rule within these outcome strata,
not sampled to estimate success rates. Selection used all three systems; this view shows
only the PPAT columns. Mesh previews are reader aids, not image inputs.
Matched test inputs do not make this a controlled backbone ablation. More examples:
\cref{sec:qualitative}, including the same cases with Uni3D-S in \cref{fig:localization-encoders}.}
\label{fig:main-localization}
\end{figure}

\subsection{Multitask capability continuation}

All localization results above continue to use R, the $31{,}434$-update
\lamselect{} checkpoint. For capability figures only, we continue the same
weights for $10{,}000$ updates with an $80/15/5$ locate/describe/point mixture.
This M10k@10k run has one seed and does not replace R in any localization table.

\paragraph{Mesh-only description.}
Neither R nor M10k@10k receives an image for this task. In a judge-assisted,
blinded review of $300$ held-out-caption objects, R already produces grounded,
non-generic mesh-only descriptions on $52.7\%$ of correct-mesh inputs. M10k@10k
reaches $63.0\%$, a paired gain of $10.3$ percentage points with a $95\%$
identity-grouped bootstrap interval of $[3.7,16.7]$. Hidden positive and negative
controls score $86.7\%$ and $5.7\%$, respectively. These are assistant judgments,
not human accuracy; protocol and controls are in \cref{sec:vlxm10k}.

\paragraph{Native pointing.}
On $2{,}000$ present and $533$ absent real-BOP queries, M10k@10k has $99.6\%$
strict-format validity, $35.7\%$ present visible-mask hit rate, and $26.6\%$
absent false positives. The present hit rate is $60.6\%$ on seen YCB-V objects;
the unseen LM-O, HB-validation, and HOPE-validation suites reach $17.8\%$,
$29.6\%$, and $34.8\%$. Thus the continuation supplies native point answers,
but unseen-object pointing and absence rejection remain weak.

Earlier \lamdirect{} capability studies and R's image-plus-mesh description and
single-prompt point probes remain in \cref{sec:capabilities}. The capability
continuation does not change the reported localization model.
\FloatBarrier

\subsection{Query color ablation}
\label{sec:ablations}
\label{sec:iso1}

ISO-1 isolates query-surface color by setting every point's RGB to gray. The two runs share
code, configuration, training steps, frozen encoder, and sampled geometry; only query color
changes. They use the earlier \lamdirect in-graph stack, so this intervention does not
estimate the color effect in \lamselect.

Removing color reduces mAP by $79.6\%$ on LM-O, $57.8\%$ on HOPE, and $27.2\%$ on HB
(\cref{tab:iso1}). This matched intervention supports a role for surface color under
the tested encoder and recipe. One run per arm does
not estimate the effect across seeds. This pair has no official BOP YCB-V evaluation.
\Cref{fig:qualappearance} illustrates the intervention on two similarly shaped statues.

\subsection{Proposal, presence, and identity errors}
\label{sec:analysis}
\label{sec:errorbudget}

We audit native \lamselect predictions on HB's $276$ images from $12$ non-development scenes:
$1{,}481$ present and $2{,}208$ absent queries (\cref{sec:protocol}), using PPAT.
Every saved selected box maps to its proposal, with zero invalid decodes.
\Cref{fig:main-localization} compares \lamselect and \lamdirect on
matched inputs. The complete encoder comparison is in \cref{tab:vlx-errors-encoders}.

\paragraph{Proposal coverage is not sufficient.}
At IoU $0.5$, let $G$ be the number of target instances, $C$ the maximum one-to-one
matching to available proposals, and $S$ the corresponding matching to selected boxes.
\Cref{tab:vlx-errors} partitions $G$ into matched instances, unavailable proposal capacity
($G-C$), and the remaining covered capacity ($C-S$), split by refusal or nonempty selection.
Most missed targets have usable proposals. This is an IoU-specific recall accounting,
not an AP decomposition or a causal diagnosis: a valid refusal alone cannot distinguish
failure to recognize the target from a decision not to commit.

\begin{table}[!htbp]
\centering\small\setlength{\tabcolsep}{6pt}
\caption{\textbf{\lamselect with PPAT on non-development HB.} Native visible-box outputs
at IoU $\ge0.5$, with maximum one-to-one matching. The first four rows partition
$1{,}481$ target instances; absent false positives use a separate denominator of $2{,}208$
queries. This run has zero invalid outputs and no multi-box answers on this cohort.
Lower counts are better for errors, higher for matched instances.}
\label{tab:vlx-errors}
\begin{tabular}{@{}lr@{}}
\toprule
Outcome & Instances / queries \\
\midrule
Matched target instances & 747 \\
No matching proposal & 68 \\
Covered, but refused & 470 \\
Covered, but selection missed & 196 \\
\midrule
Absent-query false positives & 311 \\
\bottomrule
\end{tabular}
\end{table}

\paragraph{Wrong-object selections.}
Among covered queries with no selected target hit,
$130/196$ PPAT outputs overlap another annotated object at IoU $\ge0.5$.
These are annotation-level wrong-object matches, not semantic identity judgments.
The corresponding diagnostics for HOPE and LM-O are in \cref{sec:vlxerrors-extra}.

\paragraph{Mesh substitution does not reliably redirect grounding.}
On a fixed query-swap diagnostic, both co-visible targets are followed correctly in only
$1/18$ and $5/18$ pairs for \lamdirect with PPAT and \lamselect with PPAT,
respectively. Each pair keeps the image and instruction fixed; \lamselect also keeps its
proposal set fixed. Many original source queries already fail, and these controls remain
in the denominator. Changing the answer is insufficient: both answers must hit the
corresponding target without hitting the other. Uni3D-S, exact-set, and absent-donor controls
are reported in \cref{sec:meshbinding}. These selected cases expose unreliable paired
grounding despite proposal coverage, not a population failure rate or a causal diagnosis
of every miss.

\begin{samepage}
\section{Conclusion}
\label{sec:conclusion}

\method{} uses native mesh tokens to condition two grounding VLMs. With PPAT tokens,
\lamselect selects image-region proposals and improves house-protocol and official YCB-V accuracy
over direct-coordinate \lamdirect. Paired mesh substitutions expose unreliable
target following even with proposal coverage. R already generates mesh-only
descriptions; a single-seed multitask continuation raises judge-assisted grounding
from $52.7\%$ to $63.0\%$. Its $35.7\%$ present-mask point hit rate and $26.6\%$
absent false-positive rate leave native pointing unreliable. These capability
results do not replace R as the reported localization model.

\end{samepage}

\bibliography{refs}
\bibliographystyle{plainnat}

\appendix
\section{Implementation, Data, and Evaluation}
\label{sec:implementation}
\suppressfloats[t]

\lamdirect{} and \lamselect{} adapt the complete pretrained LocateAnything-3B
\citep{wang2026locateanything} and VLX-Seek-1.5-10B \citep{omailab2026vlxseek} systems,
respectively. Earlier experiments retain their checkpoint-specific settings.

\subsection{\lamselect{} implementation}
\label{sec:vlximplementation}

VLX-Seek-1.5-10B combines a Qwen3.5 decoder, its primary vision tower, and an auxiliary
C-RADIO tower; native region encoding uses both towers. Our connector is LayerNorm$(d)$,
Linear$(d,4096)$, GELU, Linear$(4096,4096)$, with $d=512$ for PPAT and $d=1024$ for Uni3D-S.
The wider Uni3D connector adds $2{,}098{,}176$ parameters. Pretrained weights and the added
mesh-placeholder vocabulary row are frozen; projected mesh features replace its input
embeddings. \Cref{tab:hparams} gives decoder dimensions and LoRA settings.

The PPAT reference and Uni3D control share those settings and the same prepared query order.
Both use the intermediate $31{,}434$-update checkpoint, before the scheduled training horizon.

\paragraph{Targets and inference.}
The prepared localization task requests all instances and assigns targets to ground-truth
visible-box candidates. Prompt labels are masked; answer-token cross entropy includes the
terminal token. The answer format in \cref{tab:tokens} ends with \texttt{<|im\_end|>}.
Inference replaces training candidates with raster-ordered WeDetect
proposals and uses no ground truth. The reported optimized route shares image, region, and
prefix computation across groups of up to $21$ mesh queries, with fused kernels and LoRA
merged in memory. Box confidence is the FP32 softmax probability of the selected region-ID
token at its autoregressive generation step.

\begin{table}[tb]
\centering\small
\caption{\textbf{\lamselect{} prepared training manifest.} Weighted query rows sampled with
replacement. \lamselect{}'s PPAT reference and Uni3D control consume the same order; the reference checkpoint consumes only
part of the manifest. These counts are not unique-image or unique-mesh counts and do not
describe \lamdirect{}'s separately sampled training recipe.}
\label{tab:vlxdata}
\begin{tabular}{@{}lr@{}}
\toprule
Source & Prepared query rows \\
\midrule
GSO v2 & 1{,}106{,}785 \\
ABO & 978{,}042 \\
YCB-V PBR & 703{,}878 \\
Objaverse PPAT union & 627{,}079 \\
YCB-V real & 622{,}876 \\
Objaverse tabletop v2tex & 73{,}784 \\
Objaverse tabletop wave2 & 73{,}340 \\
\midrule
Total & 4{,}185{,}784 \\
\bottomrule
\end{tabular}
\end{table}

\subsection{\lamdirect{} implementation}

\Cref{tab:hparams,tab:tokens} compare the two PPAT systems. The details below describe
\lamdirect{} and its earlier ablation regimes. Its learning rates scale
with the square root of the batch-size ratio; neither system uses vision LoRA.

\paragraph{Training hardware and software.}
We use NVIDIA A100-SXM4-80GB GPUs with the NVIDIA PyTorch 25.06 container, PyTorch 2.8.0a0,
Transformers 4.57.1, PEFT 0.12.0, and DeepSpeed 0.15.4. Training uses ZeRO-1 optimizer sharding.
AdamW uses $\beta=(0.9,0.999)$ and $\epsilon=10^{-8}$.
Online packing uses a buffer of $32$ samples; images are resized aspect-preservingly
to a $1{,}280$-pixel long edge. Other training settings are in \cref{tab:hparams}.

\subsection{\lamdirect{} mesh encoding and decoding}
\label{sec:meshpath}

\paragraph{Point sampling and color.}
The surface sampler is seeded by the mesh identifier's CRC32, keeping sampled geometry fixed
across arms. Each point receives the mean vertex color of its face; missing or constant colors
use gray $0.5$. ISO-1 applies this gray substitution with the geometry sample held fixed
(\cref{sec:iso1}).

\paragraph{Encoder and connector.}
PPAT-B32, the RGB PointBERT variant of OpenShape \citep{liu2023openshape,yu2022pointbert}, returns
the full pre-pool sequence with its CLIP-alignment head \citep{radford2021clip} removed.
Set abstraction to $64$ centroids and a $12$-layer, width-$512$ transformer map
$[\mathbf{x}\,\|\,\mathbf{c}]\in\mathbb{R}^{10000\times6}$ to
$\tokens\in\mathbb{R}^{65\times512}$. The token-wise connector is LayerNorm,
$512\rightarrow2048$, GELU, then $2048\rightarrow2048$.
Uni3D \citep{zhou2024uni3d} emits $65$ width-$1024$ tokens; its connector's parameter count is
matched within $0.026\%$. The prompt order is image, instruction, mesh span, then assistant turn.

\paragraph{Training objective.} Training is the base next-token objective with coordinate positions
up-weighted $5\times$ and a soft-argmax auxiliary that applies smooth-$L_1$ and GIoU to
$\mathbb{E}[c] = \sum_{i} i \cdot p_i$ read off the coordinate logits (loss weights in
\cref{tab:hparams}).

\paragraph{Attention, decoding, and score.}
Prompt attention is causal. A six-position spatial prediction block attends to the prompt and
within the block, predicting coordinates in parallel. For multiple instances, decoding continues
after each \texttt{</box>} until termination or the configured box limit. The main benchmark
AP and runtimes use a limit of one box. We append \texttt{<box>} to the prefix for present and
absent queries alike; the model can still emit \texttt{none}. A box's AP score is the mean maximum
token probability across its four coordinate positions.

\begin{table}[tb]
\centering
\caption{\textbf{Token contracts for the two PPAT systems.} Both use $65$ frozen PPAT
features, projected into their decoder width. Added vocabulary embeddings are frozen;
projected features replace the mesh placeholders. \lamselect{} selects proposed regions;
\lamdirect{} predicts coordinates. Point and selection templates listed below are
\lamdirect{} interfaces, not measured \lamselect{} capabilities.}
\label{tab:tokens}
\footnotesize
\setlength{\tabcolsep}{4pt}
\begin{tabular}{@{}p{0.82in}>{\raggedright\arraybackslash}p{2.57in}>{\raggedright\arraybackslash}p{2.57in}@{}}
\toprule
 & \lamselect{} + PPAT & \lamdirect{} + PPAT \\
\midrule
Added tokens & \texttt{<MESH\_CONTEXT>} (id 248182) & \texttt{<MESH\_CONTEXT>}, \texttt{<mesh>}, \texttt{</mesh>} (ids 152681--152683) \\
Mesh span & $65$ \texttt{<MESH\_CONTEXT>} positions & \texttt{<mesh>}, $65$ \texttt{<MESH\_CONTEXT>} positions, \texttt{</mesh>} \\
Visual input & Native image tokens and one \texttt{<objN><objfeat>} pair per candidate & Inherited \texttt{<img>}, \texttt{<IMG\_CONTEXT>} tokens \\
Output & \texttt{<ground>mesh</ground>} then \texttt{<objects>} enclosing region IDs, followed by \texttt{</objects>} & \texttt{<box><x1><y1><x2><y2></box>}, integer coordinates in $[0,1000]$ \\
Multiple boxes & Multiple \texttt{<objN>} references within the object span & Concatenated \texttt{<box>...</box>} blocks; main benchmark caps output at one box \\
Absent & \texttt{None} inside the object span & \texttt{<box>none</box>}; plain-text \texttt{none}, id 6697 \\
Request & All candidate regions matching the mesh & Single-instance locate template \\
\bottomrule
\end{tabular}
\end{table}

\paragraph{\lamdirect{} prompt grammar.}
Coordinate tokens \texttt{<0>}\ldots\texttt{<1000>} (ids 151677--152677) and
\texttt{<box>}, \texttt{<ref>}, \texttt{<text\_mask>}, and \texttt{<null>} are inherited.
Point answers reuse \texttt{<box><x><y></box>}. The following templates retain the
original strings; \texttt{<mesh-k>} expands to the mesh span in \cref{tab:tokens}.
The benchmark uses the first template.
\begin{footnotesize}
\begin{verbatim}
<image-1> Locate the object that matches this 3D shape: <mesh-1>
<image-1> Locate every instance of the object that matches this 3D shape: <mesh-1>
<image-1> The marked object matches which of these shapes: {placeholders}
\end{verbatim}
\end{footnotesize}

\begin{table}[tb]
\centering
\caption{\textbf{Training settings of the two headline PPAT systems.} \lamselect{} is evaluated
at an intermediate checkpoint; \lamdirect{} uses the completed 100K arm~A.
``100K'' denotes identity scale, not update count. A query and a packed training
sequence are different exposure units; these are not compute-matched runs.
For \lamdirect{}, decoder LoRA and the connector train $124{,}982{,}272$ of
$3{,}668.3$\,M parameters ($3.4\%$).}
\label{tab:hparams}
\footnotesize
\setlength{\tabcolsep}{4pt}
\begin{tabular}{@{}p{1.37in}>{\raggedright\arraybackslash}p{2.28in}>{\raggedright\arraybackslash}p{2.31in}@{}}
\toprule
 & \lamselect{} + PPAT & \lamdirect{} + PPAT (arm A) \\
\midrule
Initialization & VLX-Seek-1.5-10B & LocateAnything-3B \\
Text decoder & Qwen3.5, $32$ layers, width $4{,}096$ & Qwen2.5, $36$ layers, width $2{,}048$ \\
Frozen mesh features & \multicolumn{2}{l}{PPAT, $65$ tokens of width $512$} \\
Connector widths & $512\rightarrow4096\rightarrow4096$ & $512\rightarrow2048\rightarrow2048$ \\
LoRA $r$ / $\alpha$ / dropout & 16 / 32 / 0 & 64 / 128 / 0.05 \\
LoRA targets & All language-model linear layers & $q,k,v,o$ + gate, up, down \\
Optimizer / decay & AdamW / 0.01 & AdamW / 0 \\
Peak learning rate & $10^{-4}$ & $10^{-4}$ \\
Warmup updates & $1{,}000$ & $916$ \\
Schedule / horizon & Linear / $523{,}223$ updates & Cosine / $45{,}823$ updates \\
Reported checkpoint & $31{,}434$ updates & $45{,}823$ updates \\
Per-GPU batch / accum. & 1 query / 1 & 1 pack / 1 \\
Global batch & 8 queries & 8 packs \\
Delivered exposure & $251{,}472$ queries & $366{,}584$ packs \\
Sequence caps & $4{,}096$ tokens/query & $8{,}192$/pack; $4{,}096$/sample \\
Precision / clip / seed & BF16 / 1.0 / 42 & BF16 / 1.0 / 42 \\
Activation checkpointing & Language layers & Non-reentrant \\
Training loss & Answer-token cross entropy & Coordinate weight 5.0; null 0.1; box-reg 1.0 \\
\bottomrule
\end{tabular}
\end{table}

\paragraph{Earlier \lamdirect{} regimes.}
The $5{,}000$/$10{,}000$/$20{,}000$-update ablations retain the \lamdirect{} optimizer
settings, with $200$ warmup updates and cosine decay over their own horizon.
Some early arms use LoRA rank $32$ rather than $64$. Eight packed sequences per update
contain approximately $54$--$57$ samples, supplied by eight A100-80GB GPUs with
accumulation $1$, or four with accumulation $2$. All use seed $42$; the data seed
is unset in early ablations and $42$ in later ablations and the 100K reference.

\subsection{Training-data composition}
\label{sec:datacomposition}

\Cref{tab:datafull} describes the \lamdirect{} ablation anchor; the sampling and task
routing below cover that pipeline and its 100K extension. \lamselect{} instead uses
the prepared query rows in \cref{tab:vlxdata}.
The \lamdirect{} 100K configuration adds a disjoint Objaverse source, expanding
from $28{,}334$ to $129{,}835$ mesh identities, plus the $21$ YCB-V models.

\begin{table}[tb]
\centering\small\setlength{\tabcolsep}{5pt}
\caption{\textbf{\lamdirect{} ablation-anchor training sources.} Rows count annotations; effective rows apply
the repeat factor. Objects are pool / usable after excluding $97$ GSO and $405$ ABO evaluation
identities ($502$ total). YCB-V supplies $21$ models in real photos and BlenderProc PBR
renders; the other four sources use Isaac Sim or tabletop renders. Unqueryable assets in these frames
serve as distractors. These inventory counts are not verified \lamselect{} pool counts; \lamselect{}'s
prepared query-row composition is reported separately in \cref{tab:vlxdata}.}
\label{tab:datafull}
\begin{tabular}{@{}lrrrr@{}}
\toprule
Source & \makecell[r]{Objects\\pool/use} & Rows & Rep. & \makecell[r]{Effective\\rows (share)} \\
\midrule
\texttt{ycbv\_train\_real} & 21/21 & 623{,}479 & 1.0 & 623{,}479 \;(14.90\%) \\
\texttt{ycbv\_train\_pbr} & 21/21 & 702{,}649 & 1.0 & 702{,}649 \;(16.79\%) \\
\texttt{gso\_v2} & 971/874 & \phantom{00}5{,}080 & 218.0 & 1{,}107{,}440 \;(26.46\%) \\
\texttt{abo} & 4{,}047/3{,}642 & \phantom{00}2{,}397 & 408.0 & \phantom{0}977{,}976 \;(23.36\%) \\
\texttt{objv2tex} & 11{,}909/11{,}909 & 109{,}043 & 3.40 & \phantom{0}371{,}022 \;\phantom{0}(8.86\%) \\
\texttt{objv\_wave2} & 11{,}909/11{,}909 & 118{,}505 & 3.40 & \phantom{0}403{,}216 \;\phantom{0}(9.63\%) \\
\midrule
\textbf{Total} & \textbf{28{,}836/28{,}334}\,(+21) & \textbf{1{,}561{,}153} & & \textbf{4{,}185{,}782} \\
\bottomrule
\end{tabular}
\end{table}

\paragraph{Sampling and task routing.}
Rendered sources use a train-time composer; YCB-V uses stored examples.
Each composed sample has $K{=}4$ query slots, independently replaced by absent queries
with probability $0.25$. A per-sample choice selects single-instance or every-instance
localization with probability $0.5/0.5$, including for arm~A's localization samples.
Every-instance targets concatenate all matching boxes.
Repeated instances occur in $85\%$ of \texttt{objv2tex} frames.

The anchor requests \texttt{locate:0.54, point:0.18, select\_dot:0.18, describe:0.10}.
Missing task stores fall back to \emph{locate}; an instrumented run rewrote $54.4\%$
of requested \emph{select} samples. Delivered fractions were not recorded for any
$10{,}000$-step run. Arm~A uses a manifest-defined per-source router without fallback
and receives no \emph{point} or \emph{describe} samples. The PPAT and Uni3D
appearance-token paths bypass feature-noise augmentation.

\paragraph{Category provenance.}
Category summaries count each eligible source/mesh ID once, including absent-query candidates.
Native labels and object names are supplemented by noun rules over Cap3D captions
\citep{luo2023cap3d}, stored Qwen mesh descriptions, and product titles.
Ambiguous entries remain unassigned. These are approximate metadata categories,
not visually verified labels or training-sampling frequencies.

\paragraph{Contamination audit.}
Synthetic-source recipes define object-level held-out splits, with load-time ID-disjointness
checks. We compare each evaluation CAD against the training pool in the frozen mesh
encoder's L2-normalized CLS space and manually review pairs above cosine similarity
$0.90$. Reported contamination counts are lower bounds. The audit flags no pairs in
LM-O, HOPE, HB, T-LESS, TUD-L, IC-BIN \citep{doumanoglou2016icbin}, or ITODD
\citep{drost2017itodd}; their maximum top-1 similarity is $0.864$.

\subsection{Mesh onboarding}
\label{sec:onboardingbench}
Mesh encoding and descriptor construction are separate from cached-token inference
(\cref{sec:blackwell-timing}). PPAT and matched CNOS component timings have not been
measured on GB200. The reported inference times exclude mesh encoding and cannot
supply an onboarding measurement.

\subsection{Evaluation protocols}
\label{sec:protocoldetails}

\subsubsection{Cohorts and sampling}
\label{sec:samplingdetail}

\Cref{tab:suites} lists the saved evaluation cohorts. Official YCB-V AP uses the
$900$ target images with all $21$ object queries per image. The Blackwell timing
pilot uses a separate eight-image subset and house scoring; its checkpoints and
timing exclusions are given in \cref{sec:blackwell-timing}.
Historical $300$-image visible-box scores cannot substitute for official BOP results.
HB samples strided frames across all scenes; LM-O retains the first frames selected
when its manifest was built.

HB $\discrim$ excludes development scene \texttt{000001}, leaving $12$ scenes with
$1{,}481$ present and $2{,}208$ absent queries. This population is distinct from
the all-scene HB mAP cohort.

\begin{table}[tb]
\centering
\caption{\textbf{Evaluation cohorts and coverage of the two PPAT systems.} A check marks
a measured result; \textit{n/e} means not evaluated in the accepted record. Shared cohort
counts alone do not establish parity; saved image/query inputs and the scorer have
been matched for HOPE, LM-O, and HB. Absent queries use up to eight sampled classes per image;
YCB-V queries all $21$ models under official BOP amodal scoring; the other rows use
the internal visible-box protocol. HB and HOPE use validation splits with public ground truth.
Simulated rows retain the original splits and the contamination limitation in \cref{sec:data}.}
\label{tab:suites}
\footnotesize
\setlength{\tabcolsep}{3pt}
\begin{tabular}{@{}llrrrrcc@{}}
\toprule
Suite & Split & Obj. & Scenes & \makecell[r]{Images\\eval.} & \makecell[r]{Present\\queries} & \makecell{\lamselect{}\\+ PPAT} & \makecell{\lamdirect{}\\+ PPAT} \\
\midrule
YCB-V BOP & official targets & 21 & 12 & 900 & 4{,}123 & \cmark & \cmark \\
LM-O & \texttt{lmo/test} & 8 & 1 & 300 & 2{,}319 & \cmark & \cmark \\
HOPE & \texttt{hope/val} & 28 & 10 & 50 & \phantom{0}775 & \cmark & \cmark \\
HB & \texttt{hb/val\_prim.} & 33 & 13 & 300 & \makecell[r]{1{,}553\\(2{,}400 abs.)} & \cmark & \cmark \\
\midrule
GSO/ABO/Objv.\ h/o & renders & \makecell[r]{97 / 405\\/ 385} & \na & 300 & \na & \notmeasured & \cmark \\
\bottomrule
\end{tabular}
\end{table}

\subsubsection{Official BOP scoring}
\label{sec:bopprotocol}

The official YCB-V target list spans $12$ test scenes. Ground truth uses mask-derived
amodal boxes; instances below $10\%$ visibility are ignored. COCO AP@$[.50{:}.05{:}.95]$
is macro-averaged over objects. Querying every object on every image yields $18{,}900$
queries, $78.2\%$ absent. Ground truth contains $4{,}125$ instances, including
$2$ ignored instances.

\lamselect{} preserves every selected box and its generation-token score;
\lamdirect{} uses its batched fast route. The latter has $26$ invalid answers;
both \lamselect{} encoder variants have none. Invalid outputs contribute no detections
and never count as valid refusals. We report each evaluation route's own AP;
small numerical changes between reruns are not an uncertainty estimate.
On GDRNPPDet\_PBRReal's published detections, the scorer returns $0.85215$ AP against
the reported $0.852$; all twelve COCO statistics agree within $10^{-3}$.

\subsubsection{Matched internal evaluation}

\lamdirect{} uses the \lamselect{} manifests for $1{,}175$ HOPE and $2{,}400$
LM-O queries and the same visible-box AP kernel, yielding mAP $0.15465$ and $0.18144$.
It has $21$ and $132$ invalid answers, respectively, all with empty decoded boxes.
Together with the matched HB query audit, this matches query selection and scoring
across systems, while architectures and output cardinalities differ.

Internal HB uses public \texttt{val\_primesense}: $13$ scenes, $33$ objects and
modal boxes. BOP-Classic HB instead uses $3$ scenes and $16$ objects from withheld
\texttt{test\_primesense}, with amodal boxes. These protocols are not interchangeable.
External comparisons use official YCB-V AP (\cref{tab:vlxencoder-bop});
missing official evaluations remain marked as unmeasured.

\subsubsection{Presence discrimination}

A \emph{commit} is a valid predicted box. At natural decoding, we measure how often a model
answers present queries relative to absent ones:
\begin{equation}
\discrim \;=\; \frac{P(\text{commit} \mid \text{present})}{P(\text{commit} \mid \text{absent})},
\label{eq:discrim}
\end{equation}
A value of $1.0$ means committing is independent of presence. A committed box can still
identify the wrong object; a hit requires target IoU $\ge0.5$.
When no absent query is answered, we report $+\infty$ for the ratio.
HB excludes development scene \texttt{000001}, as specified above.

\section{Supporting Quantitative Results}
\label{sec:extraresults}
\suppressfloats[t]

The two headline systems are \lamselect{} and \lamdirect{}, both with PPAT.
Historical experiments
below retain their original checkpoints and conditions; they do not constitute
matched ablations of both headline systems.
Official YCB-V AP uses the $900$-image BOP protocol with mask-derived amodal
boxes and visibility-ignore handling. The separate Blackwell speed pilot
(\cref{sec:blackwell-timing}) reports house mAP on eight images. Other datasets
retain the internal protocols specified in each caption.

\subsection{Grounding errors across suites}
\label{sec:vlxerrors-extra}
\label{sec:baselineerrors}

\Cref{tab:vlx-errors-encoders} retains both mesh encoders for the HB instance
accounting in \cref{sec:analysis}. They share image/query inputs and ordered proposals.
Uni3D-S recovers more present targets but emits more boxes on absent queries.
Among covered queries with no selected target hit, $130/196$ PPAT and $129/181$
Uni3D-S outputs overlap another annotated object at IoU $\ge0.5$.

\begin{table}[tb]
\centering\small\setlength{\tabcolsep}{6pt}
\caption{\textbf{Full mesh-encoder comparison of \lamselect{} HB errors.}
Native visible-box outputs on $276$ non-development
images at IoU $\ge0.5$, with maximum one-to-one matching. The first four rows
partition $1{,}481$ target instances; absent false positives use $2{,}208$ queries.
Both runs have zero invalid outputs and no multi-box answers on this cohort.
Higher is better for matched instances; lower is better for errors.}
\label{tab:vlx-errors-encoders}
\begin{tabular}{@{}lrr@{}}
\toprule
Outcome & PPAT & Uni3D-S \\
\midrule
Matched target instances & 747 & \best{826} \\
No matching proposal & 68 & 68 \\
Covered, but refused & 470 & \best{406} \\
Covered, but selection missed & 196 & \best{181} \\
\midrule
Absent-query false positives & \best{311} & 352 \\
\bottomrule
\end{tabular}
\end{table}

\Cref{tab:vlx-errors-extra} applies the same accounting to HOPE and LM-O.
HOPE includes multi-instance queries; its instance counts differ from query counts.
All native boxes are retained, and both runs have zero invalid outputs.

\begin{table}[tb]
\centering\small\setlength{\tabcolsep}{4pt}
\caption{\textbf{\lamselect{} instance and absence errors on HOPE and LM-O.} Native visible-box
outputs on $50$ and $300$ images, respectively, at IoU $0.5$. Matched, uncovered,
covered/refused and covered/missed counts sum to the number of target instances
($920$ on HOPE; $2{,}319$ on LM-O). The final column counts absent queries emitting
a box, not the number of false-positive boxes. These single-threshold counts do not decompose AP.}
\label{tab:vlx-errors-extra}
\begin{tabular}{@{}llrrrrr@{}}
\toprule
Suite & Encoder & Matched & Uncovered & \makecell{Covered\\refused} &
\makecell{Covered\\missed} & \makecell{Absent FP\\/ queries} \\
\midrule
HOPE & PPAT & \best{600} & 0 & \best{26} & 294 & \best{245}/400 \\
& Uni3D-S & 577 & 0 & 83 & \best{260} & 272/400 \\
\midrule
LM-O & PPAT & \best{1{,}414} & 249 & \best{415} & \best{241} & \best{11}/81 \\
& Uni3D-S & 1{,}241 & 249 & 528 & 301 & 25/81 \\
\bottomrule
\end{tabular}
\end{table}

The instance partition does not count extra predictions on queries with a target hit.
HOPE has multi-box outputs; LM-O and non-development HB do not.

\paragraph{Earlier \lamdirect{} wrong-object matches.}
On HB's non-development scenes, $142$ of arm~A's $163$ present-query predictions with target
IoU below $0.10$ overlap a different annotated object at IoU $\ge0.5$.
This direct-coordinate diagnostic is separate from the proposal-conditioned
accounting above.

\subsection{Paired mesh-binding controls}
\label{sec:meshbinding}

We fix $36$ distinct images before inference, $12$ each from HOPE, LM-O, and
non-development HB. Each suite contributes six co-visible and six absent donors.
Half of each donor group uses scene-balanced uniform selection; half uses
feature-neighbor proxies, not visually verified lookalikes. All source and co-visible
donor instances have proposal coverage at IoU $0.5$, and source--donor GT-box IoU is
below $0.1$. This is a selected diagnostic favorable to proposal selection.

Each pair changes only the valid mesh tokens and their cache identity, holding the image,
native generic instruction, decoder settings, batch size one, and batch position fixed.
\lamselect{} receives identical ordered proposals in both conditions; \lamdirect{} remains
proposal-free. Native prompts and output cardinalities differ between architectures.

\begin{table}[tb]
\centering\small\setlength{\tabcolsep}{5pt}
\caption{\textbf{Paired target following under mesh substitution.} Success counts out of
$18$ pairs in each donor group, at IoU $0.5$. Co-visible success requires a source hit
without a donor hit before substitution, and a donor hit without a source hit afterward.
Absent-pair success requires a source hit followed by a valid empty answer.
Primary success allows unrelated extra boxes; exact-set success requires a one-to-one
match for every target instance and no extras in both conditions. Failed source controls and invalid answers
remain in the denominators.}
\label{tab:meshbinding}
\begin{tabular}{@{}lrrrr@{}}
\toprule
& \multicolumn{2}{c}{Co-visible donor / 18} & \multicolumn{2}{c}{Absent donor / 18} \\
\cmidrule(lr){2-3}\cmidrule(l){4-5}
System & Primary & Exact set & Primary & Exact set \\
\midrule
\lamdirect{} + PPAT & 1 & 0 & 9 & \best{9} \\
\lamselect{} + PPAT & \best{5} & \best{3} & \best{11} & \best{9} \\
\lamselect{} + Uni3D-S & 4 & \best{3} & 10 & \best{9} \\
\bottomrule
\end{tabular}
\end{table}

Source controls fail on $14/18$, $10/18$, and $10/18$ co-visible pairs in table order;
thus paired failures cannot all be attributed to substitution. Answer sets change on
$14/18$, $14/18$, and $12/18$ pairs, far more often than both targets are followed
correctly. \lamdirect{} has two invalid condition outputs among $72$; both \lamselect{} runs
have none. Invalid answers never count as valid rejection. Four co-visible and two
absent pairs contain multiple target instances, which prevents exact-set success under
\lamdirect{}'s native one-box limit. These small, selected counts do not establish
a reliable model ranking or isolate the backbone's effect.

\subsection{\lamdirect{} synthetic transfer}
\label{sec:100kconfiguration}

\begin{table}[tb]
\centering\small\setlength{\tabcolsep}{6pt}
\caption{\textbf{Synthetic-suite transfer of the selected \lamdirect{} checkpoint.}
Arm~A at $45{,}823$ updates, with PPAT and one-box decoding.
House visible-box mAP on $300$ strided images per held-out suite, with
\texttt{max\_neg\_per\_image}${=}8$. These use the original held-out splits before
duplicate removal (\cref{sec:datacomposition}).
This single configuration does not measure the effect of training-set scale.}
\label{tab:scale100k}
\begin{tabular}{@{}lrrr@{}}
\toprule
System & GSO & ABO & Objaverse \\
\midrule
\lamdirect{} + PPAT & 0.305 & 0.353 & 0.320 \\
\bottomrule
\end{tabular}
\end{table}

\subsection{\lamselect{} mesh-encoder comparison}
\label{sec:vlxencoders}

PPAT remains the primary mesh encoder; \cref{tab:vlxencoder,tab:vlxencoder-bop}
report the secondary Uni3D-S comparison. The two models share training-query order,
seed, optimizer settings and schedule (\cref{sec:vlximplementation}).
This compares each encoder with its required connector.
One training trajectory per encoder does not establish a statistically significant ordering.
PPAT leads on HOPE, LM-O and YCB-V, while Uni3D-S leads on HB but emits more
absent-query false positives (\cref{tab:vlx-errors-encoders}). HOPE top-one AP is
$31.90\%$ for PPAT and $26.81\%$ for Uni3D-S; LM-O and HB AP are unchanged by
the top-one restriction.

\begin{table}[tb]
\centering\small\setlength{\tabcolsep}{3.5pt}
\caption{\textbf{Official BOP YCB-V encoder accuracy and reported Blackwell time.}
All $900$ target images, $12$ scenes and $18{,}900$ image--object queries;
mask-amodal COCO AP@[.50:.05:.95] with BOP low-visibility ignore handling.
\lamselect{} checkpoints have $31{,}434$ updates; \lamdirect{} arm A has $45{,}823$.
All systems use the same official target list and scoring protocol for AP.
Times are approximate median cohort-mean seconds/image from three repeats of
the separate eight-image GB200 batch-$21$ pilot; ranges span those repeats.
The \lamdirect{} timing checkpoint has $10{,}000$ updates rather than $45{,}823$;
\lamselect{} timing uses cached proposals and excludes live detection.
These are assembled accuracy/runtime summaries; see \cref{sec:blackwell-timing}.}
\label{tab:vlxencoder-bop}
\begin{tabular}{@{}llrrrrl@{}}
\toprule
System & Encoder & AP & AP50 & AP75 & \makecell{GB200 s/img\\$\downarrow$} & \makecell{Repeat range\\s/img} \\
\midrule
\lamdirect{} & PPAT & 0.6384 & 0.7885 & 0.7112 & \best{0.322} & 0.309--0.324 \\
\lamselect{} & PPAT & \best{0.7152} & \best{0.9150} & \best{0.8183} & 0.682 & 0.652--0.837 \\
\lamselect{} & Uni3D-S & 0.7134 & 0.9110 & 0.8162 & 0.882 & 0.705--0.921 \\
\bottomrule
\end{tabular}
\end{table}

\begin{table}[tb]
\centering\small\setlength{\tabcolsep}{5pt}
\caption{\textbf{Secondary mesh-encoder comparison within \lamselect{}.} Native house visible-box mAP
at $31{,}434$ updates on $50$ HOPE images and $300$ images each for LM-O and HB.
Both encoders share queries and detector proposals. HB's
development-excluded column uses $276$ images; the full HB column includes all $13$ scenes.
Official YCB-V scores and the separate Blackwell timings are in \cref{tab:vlxencoder-bop}. These are encoder-plus-connector
comparisons, separate from the two-system comparison in \cref{tab:vlxmain}.}
\label{tab:vlxencoder}
\begin{tabular}{@{}lrrrr@{}}
\toprule
Mesh encoder & HOPE & LM-O & HB (all) & HB (no dev.) \\
\midrule
PPAT & \best{0.3707} & \best{0.3880} & 0.3658 & 0.3404 \\
Uni3D-S & 0.3160 & 0.3337 & \best{0.4559} & \best{0.4219} \\
\bottomrule
\end{tabular}
\end{table}

\subsection{\lamdirect{} multiple-instance localization}
\label{sec:multibox}

Arm~A receives the same image--mesh inputs under three prompt/cap settings (\cref{tab:multibox}).
The HOPE panel is fixed before inference, spans ten scenes and shares views; it is not a
representative benchmark. Present targets have at least $50\%$ visibility and visible boxes covering
at least $0.5\%$ of the image. All eligible targets are matched one-to-one at IoU $\ge0.5$.

\begin{table}[tb]
\centering\small\setlength{\tabcolsep}{4pt}
\caption{\textbf{\lamdirect{} prompt and box-limit comparison on selected HOPE queries.}
On $48$ two-instance queries, recall (\%) uses $96$ targets and precision (\%) uses all valid
predicted boxes. Both-found counts require well-formed answers but allow extras. Single exact
means one matched box with no extras; absent FP counts queries emitting a box (eight controls each).
\lamselect{}'s native multi-box benchmark results use a different query population and cannot fill this pilot.}
\label{tab:multibox}
\begin{tabular}{@{}lrrrrrr@{}}
\toprule
Prompt & Cap & Recall & Precision & \makecell{Both\\found / 48} &
\makecell{Single\\exact / 8} & \makecell{Absent\\FP / 8} \\
\midrule
Single & 1 & 24.0 & \best{48.9} & 0 & \best{4} & 7 \\
Single & 4 & 29.2 & 21.9 & 4 & 2 & 7 \\
Every & 4 & \best{41.7} & 22.2 & \best{10} & 0 & 7 \\
\bottomrule
\end{tabular}
\end{table}

\begin{figure}[tb]
\centering
\includegraphics[width=\textwidth]{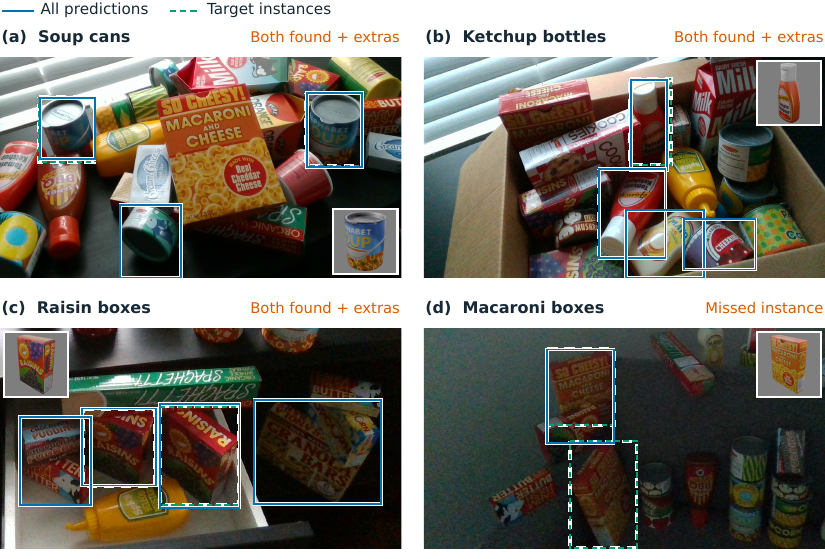}
\caption{\textbf{\lamdirect{} multiple-instance localization on real HOPE images.} Every-instance prompt,
four-box limit. Three selected examples find both targets with extra boxes; the fourth misses
one target. Blue boxes show every prediction, dashed green boxes show all targets, and insets
show the queried mesh. Display crops retain all boxes.}
\label{fig:multibox-qualitative}
\end{figure}

No two-instance query yields an exact-set-correct answer in any mode. Larger caps recover
more instances but add false positives (\cref{fig:multibox-qualitative}). Box precision and
recall retain valid boxes from two malformed every-instance answers; these contribute three
matched and four extra boxes, but cannot count as well-formed answers.

\subsection{Blackwell batch scaling}
\label{sec:blackwell-timing}

The GB200 pilot evaluates eight YCB-V images with $168$ queries ($35$ present,
$133$ absent), after $21$ excluded warmup queries on a separate image.
\Cref{tab:blackwell-timing,tab:blackwell-batches} report three fresh-process repeats per configuration.
The \lamdirect encoder pair has $10{,}000$ updates, unlike the $45{,}823$-update
PPAT accuracy checkpoint; the \lamselect pair has $31{,}434$ updates in both
evaluations. The pilot's timing cohort differs from official $900$-image AP.

\begin{table}[!htbp]
\centering\small\setlength{\tabcolsep}{4pt}
\caption{\textbf{Reported Blackwell GB200 timing.} Eight images, $21$ queries per image,
three repeats. Throughput is the median across repeats. Approximate batch-$21$
seconds/image is $21$ divided by the reported throughput, the median of the repeat-level
cohort means, not the median of individual-image latencies. Peak allocated memory
covers the reported batch sweep, including loading and warmup. Baseline timings retain
malformed answers: $2/168$ queries for PPAT and $1/168$ for Uni3D-S per repeat.
Full batch results and timer boundaries are in \cref{sec:blackwell-timing}.}
\label{tab:blackwell-timing}
\begin{tabular}{@{}llrrrr@{}}
\toprule
System & Encoder & \makecell{Queries/s\\$b=1$} & \makecell{Queries/s\\$b=21$}
& \makecell{Approx. s/img\\$b=21$} & \makecell{Peak alloc.\\GiB} \\
\midrule
\lamdirect{} & PPAT & \best{7.89} & \best{65.3} & \best{0.322} & 7.9--8.5 \\
\lamdirect{} & Uni3D-S & 7.70 & 59.9 & 0.351 & 7.9--8.4 \\
\lamselect{} & PPAT & 2.07 & 30.8 & 0.682 & 21.4 \\
\lamselect{} & Uni3D-S & 1.57 & 23.8 & 0.882 & 21.4 \\
\bottomrule
\end{tabular}
\end{table}

\begin{table}[!htbp]
\centering\small\setlength{\tabcolsep}{4pt}
\caption{\textbf{Reported GB200 throughput by query-batch size.} Median queries/s
across three repeats on the same eight-image cohort. Batch size counts mesh queries
sharing one image. The final column is the observed minimum--maximum over the three
batch-$21$ repeats; ranges for other batches were not supplied. Malformed baseline
answers remain in the query count.}
\label{tab:blackwell-batches}
\begin{tabular}{@{}llrrrrrl@{}}
\toprule
System & Encoder & $b=1$ & $b=4$ & $b=8$ & $b=16$ & $b=21$ & $b=21$ range \\
\midrule
\lamdirect{} & PPAT & \best{7.89} & \best{21.5} & \best{35.8} & 44.2 & \best{65.3} & 64.9--67.9 \\
\lamdirect{} & Uni3D-S & 7.70 & 20.4 & 33.9 & \best{44.4} & 59.9 & 52.7--63.4 \\
\lamselect{} & PPAT & 2.07 & 6.86 & 12.5 & 16.5 & 30.8 & 25.1--32.2 \\
\lamselect{} & Uni3D-S & 1.57 & 5.45 & 9.9 & 14.1 & 23.8 & 22.8--29.8 \\
\bottomrule
\end{tabular}
\end{table}

\paragraph{Timing scope.}
CUDA-synchronized timing includes shared image processing, vision and prefix work,
and grouped query inference; it excludes image/token file reads, loading, warmup,
scoring, mesh encoding and live proposal detection. \lamselect uses $100$ cached
proposals per image; \lamdirect uses detector-free one-box decoding.
The reported environment is NGC PyTorch~26.01, PyTorch~2.10 and CUDA~13.1 on GB200.
The benchmark uses GPU~0 while light matrix multiplication occupies GPUs~1--3;
shared power and CPU activity may affect timing. Peak allocated memory includes
loading and warmup. Approximate seconds/image is $21$/median queries/s, the median
of the three repeat-level cohort means, not individual-image latency. Repeat ranges
are observed minima and maxima, not confidence intervals.

\paragraph{Quality accounting.}
Reported house mAP is $0.832$ for \lamselect PPAT and $0.826$ for Uni3D-S at
every batch size. \lamdirect ranges across batches are $0.717$--$0.722$ and
$0.677$--$0.681$; the report supplies no exact batch-$21$ values. Its PPAT model
returns two malformed absent-query answers per $168$ queries; Uni3D-S returns one.
These remain in the timing denominator and do not count as valid abstentions.
The small cohort, training budgets and decoder paths limit this pilot
to batching and execution cost.

\subsection{Leaderboard comparison scope}
\label{sec:readingfig2}

\Cref{fig:pareto} shows points from the
2026-09-06 snapshot of BOP's
\href{https://bop.felk.cvut.cz/leaderboards/detection-bop22/ycb-v/}{seen-object}
and \href{https://bop.felk.cvut.cz/leaderboards/detection-unseen-bop23/ycb-v/}{unseen-object}
2D detection boards. The separate external-only frontiers use $10$ timed
seen-object entries and $18$ named unseen-object entries. Shaded regions show
the area dominated by each external frontier.
Our models use YCB-V training identities and remain excluded from both frontiers.
The shared test images and AP definition do not equalize training exposure or
establish eligibility for the unseen-object task.

\paragraph{External timing and exclusions.}
BOP's published \emph{Time} is the average time to predict all objects in an
image. Our GB200 values cover all $21$ YCB-V queries with the scope in
\cref{sec:blackwell-timing}. External runtimes retain their original hardware and
timing scopes without normalization; the plot does not support controlled speed claims.
The submission named only ``\textsf{.}'' is excluded because it has no identifiable
method name, paper or code. Named unpublished entries with positive reported time
remain. Three seen-object entries omit runtime and therefore have no cost-axis
coordinate: DLZDet-PBR1, Extended FCOS-PBR (v2), and DLZDet-PBR+Real.
One-time mesh onboarding is excluded from these image/query times
(\cref{sec:onboardingbench}).

\begin{table}[tb]
\centering\small\setlength{\tabcolsep}{3.5pt}
\caption{\textbf{\lamdirect{} ISO-1 color and gray twins.} Identical configuration, LoRA $r{=}64$, and
$10{,}000$ updates; token stores differ only in \texttt{--force-gray}. HB mAP uses strided-300
over 13 scenes; $\discrim$ uses scenes 2--13. Both arms re-score
bit-identically; training-run uncertainty remains (\cref{sec:protocol}). Both checkpoints use
the retired in-graph stack and are compared with each other. Neither headline checkpoint
has a matched color/gray training pair under this protocol.}
\label{tab:iso1}
\begin{tabular}{lccccccc}
\toprule
Query point cloud & LM-O & HOPE & HB & $\discrim$ & GSO & ABO & Objv. \\
\midrule
Color twin & \best{0.198} & \best{0.122} & \best{0.244} & \best{2.356} & \best{0.266} & \best{0.302} & \best{0.257} \\
Gray twin & 0.040 & 0.052 & 0.177 & 1.798 & 0.124 & 0.241 & 0.208 \\
\midrule
$\Delta$ gray vs.\ color & $-79.6\%$ & $-57.8\%$ & $-27.2\%$ & $-23.7\%$ & $-53.3\%$ & $-20.2\%$ & $-19.4\%$ \\
\bottomrule
\end{tabular}
\end{table}

\begin{table}[tb]
\centering\small
\caption{\textbf{\lamdirect{} frozen encoder comparison.} House visible-box mAP.
The historical Uni3D variants have no official YCB-V evaluation;
headline-system scores appear in \cref{tab:vlxencoder-bop}.
All arms use $10{,}000$ updates, seed 42, and
LoRA $r{=}64$. The first two change only the encoder package; Uni3D-G also changes canonical
axis, dtype, and topology. HB commit, FP, and $\discrim$ use scenes 2--13; HB mAP uses all 13.
\lamselect{}'s separate encoder experiment uses a different schedule (\cref{tab:vlxencoder}).}
\label{tab:encoder}
\begin{tabular}{lcccccc}
\toprule
Encoder package & HB & LM-O & HOPE & commit & FP$_{\text{abs}}$ & $\discrim$ \\
\midrule
PPAT-B32 (anchor) & 0.282 & 0.140 & 0.117 & 0.637 & 0.210 & \best{3.030} \\
Uni3D-S           & 0.265 & \best{0.271} & 0.150 & 0.700 & 0.271 & 2.585 \\
Uni3D-G (Y-up)    & \best{0.304} & 0.139 & \best{0.166} & 0.603 & 0.208 & 2.901 \\
\bottomrule
\end{tabular}
\end{table}

\section{Additional Qualitative Results}
\label{sec:qualitative}
\suppressfloats[t]

\begin{figure}[p]
\centering
\includegraphics[width=0.9\textwidth]{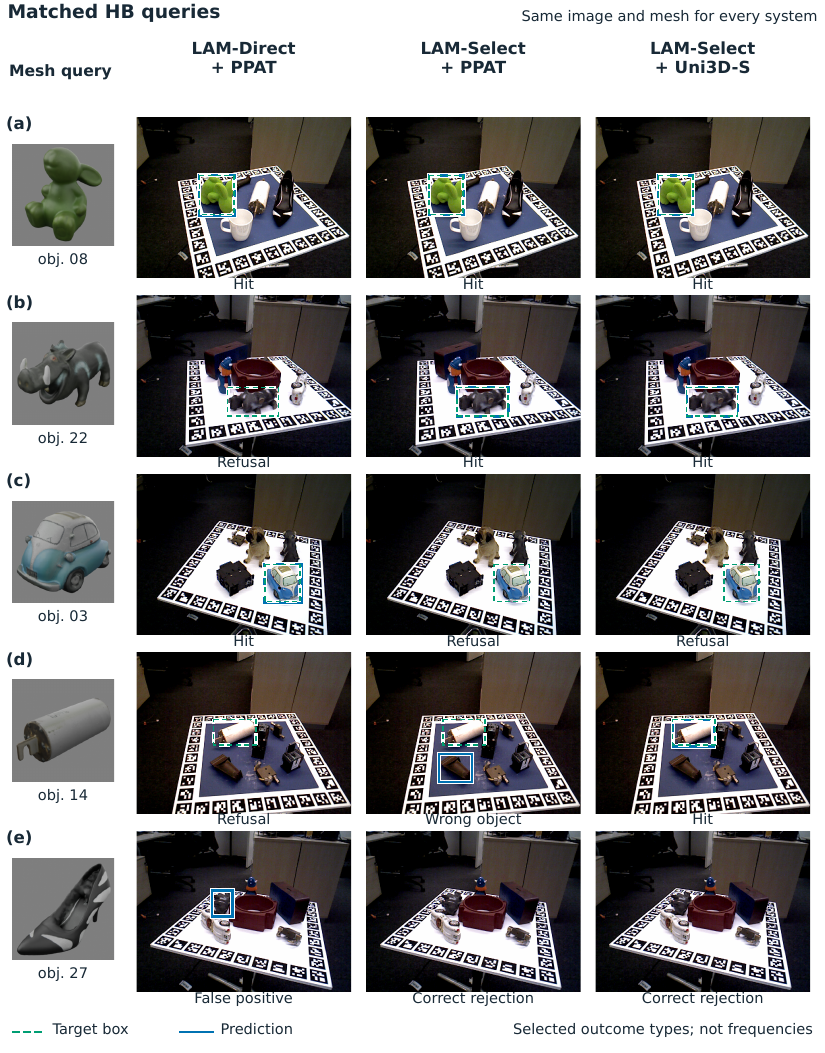}
\caption{\textbf{Matched HB detections with the full encoder comparison.}
\lamdirect{} with PPAT and \lamselect{}
with PPAT or Uni3D-S receive identical image/query inputs.
Blue boxes are predictions; dashed green boxes are visible targets. Hits require
IoU $\ge0.5$. All predictions are shown. These are the same five selected cases
as \cref{fig:main-localization}; selection used all three systems, including the
Uni3D-S control. Mesh previews are reader aids, not image inputs.}
\label{fig:localization-encoders}
\end{figure}

\paragraph{Matched systems and encoder control.}
\Cref{fig:localization-encoders} adds the Uni3D-S control to the main gallery's
PPAT pair. Selection joins saved outputs by image and mesh ID, checks identical
ground truth, and excludes HB's development scene. Five prespecified strata cover
shared success, both \lamselect{} variants succeeding where \lamdirect{} fails,
the reverse, a verified \lamselect{} wrong-object selection, and absent-query
disagreement. Present targets occupy $2.5$--$40\%$ of the image; we select the
lower-median target area within each stratum, breaking ties by query ID. The absent
stratum uses the lower-median query ID. Wrong-object matches require target IoU
below $0.1$ and overlap of at least $0.5$ with another annotated object.
These balanced strata do not estimate outcome frequencies.

Other panels use earlier \lamdirect{} dumps, with literal outputs and pixel boxes.

\subsection{Outcomes across suites}

\begin{figure}[p]
\centering
\captionsetup{font=footnotesize}
\includegraphics[width=\textwidth]{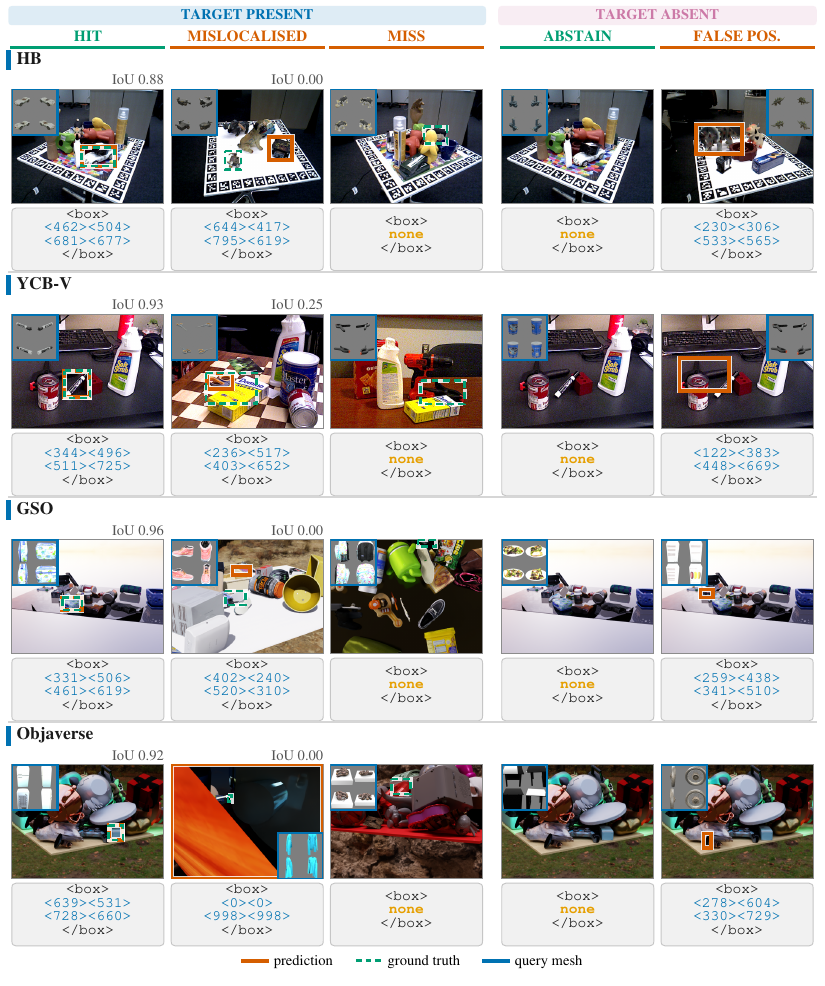}
\caption{\textbf{Localization across datasets.} Solid orange boxes show predictions; dashed
green boxes show ground truth. Insets show query meshes, and text below each image gives the
model output. Hits have IoU$\ge0.5$.}
\label{fig:qualdatasets}
\end{figure}

\Cref{fig:qualdatasets} uses \lamdirect{} 100K arm~A, with strided-$300$ dumps for
GSO and Objaverse. Hits and mislocalizations use median IoU; misses use median
relative target area. Absent-query examples use the lowest object ID in the hit's
image when available, otherwise the median image--object pair.

HB annotations permit the wrong-object checks in \cref{sec:baselineerrors}; low
target IoU alone does not verify such a match. Synthetic scenes annotate only
$1.2$--$2.2$ of roughly $25$ objects per frame, precluding the same comparison.

\subsection{Color and object identity}

\begin{figure}[p]
\centering
\captionsetup{font=footnotesize}
\includegraphics[width=\textwidth]{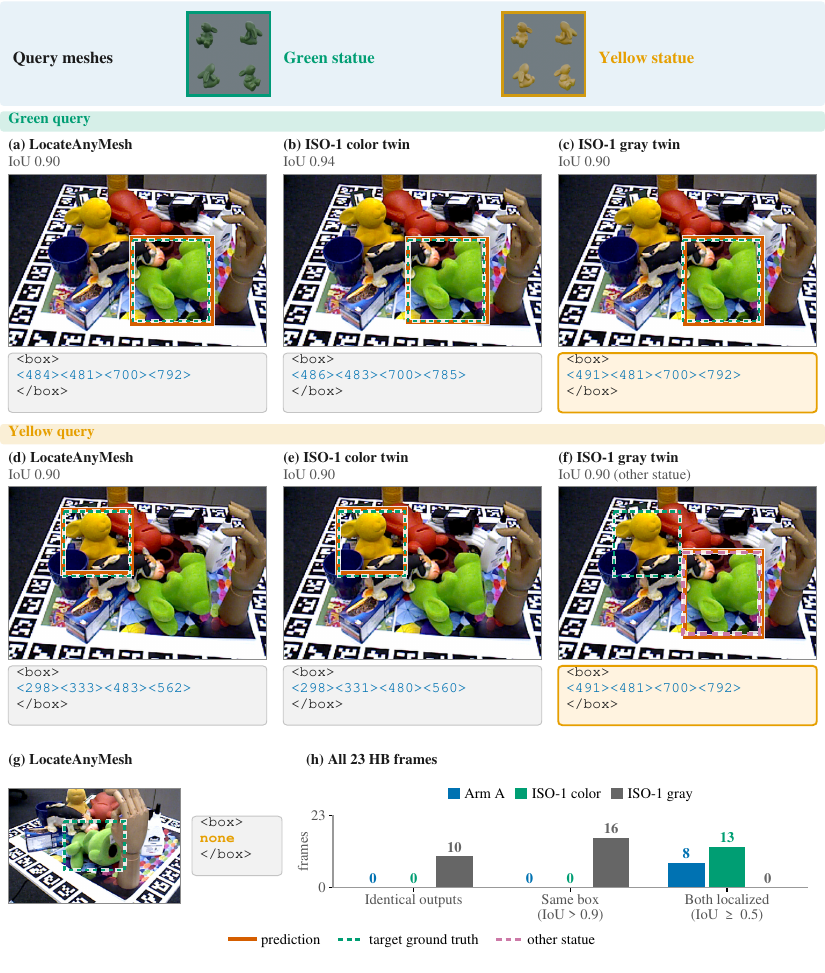}
\caption{\textbf{Query color separates similar statues.} (a--f) Two mesh queries in one HB image;
the gray twin gives identical outputs in (c,f). Only the ISO-1 twins isolate query color;
arm~A is shown for context. (g) A refusal in another frame. (h) All $23$ frames containing both statues.}
\label{fig:qualappearance}
\end{figure}

For \cref{fig:qualappearance}, we select the closest of $528$ HB CAD pairs by sorted
bounding-box extents and a scale-normalized radial descriptor before inspecting
predictions. The frame is outcome-selected from $23$ co-present frames: it maximizes
the minimum of four IoUs, requiring identical gray-twin outputs and both other models
to localize both queries at IoU$\ge0.5$. Across these frames, the gray twin never
localizes both queries correctly; \lamdirect{} arm~A does so on $8$.

\subsection{Pointing and description}
\label{sec:capabilities}
\suppressfloats[t]

The \lamdirect{} studies use a separate $20{,}000$-step multitask checkpoint with
PPAT and rank-$64$ LoRA, not either headline localization checkpoint. A later
\lamselect{} continuation supplies the main capability figures
(\cref{sec:vlxm10k}). Historical tests of R, the reported localization model,
cover native point-prompt transfer and image-plus-mesh description
(\cref{sec:vlxpoint,sec:vlxdescription}).

\paragraph{Mesh-only descriptions.}
With no image input, \texttt{Describe this 3D shape: <mesh-1>} elicits paired
answers for $120$ held-out IDs ($40$ each from ABO, GSO, and Objaverse), using the
correct mesh or a random same-source swap. Greedy decoding has a $256$-token
limit. All $240$ outputs are valid and nonempty; every pair differs. IDs are
disjoint from training lists, but geometric deduplication is unverified and
swaps may change category, so fine-grained identity is not isolated.

A blinded \texttt{gpt-6-astra} judge scores agreement with stored
Qwen3-VL-8B-Instruct captions (\cref{tab:tdm-description}). Correct meshes raise
same-asset judgments by $9.17$ percentage points ($95\%$ paired bootstrap interval
$[4.17,14.17]$), below the predeclared $20$-point success threshold.
Reference captions can misstate geometry, so agreement is not visual correctness.

\begin{table}[tb]
\centering
\small
\begin{tabular}{lrrrr}
\toprule
& \multicolumn{2}{c}{Agreement rating / 10} & \multicolumn{2}{c}{Same-asset judgments} \\
System / source & Correct mesh & Random swap & Correct mesh & Random swap \\
\midrule
\multicolumn{5}{l}{\lamdirect{} + PPAT, separate multitask checkpoint} \\
\midrule
ABO ($n=40$) & 3.73 & 1.18 & 7/40 & 0/40 \\
GSO ($n=40$) & 3.63 & 1.30 & 5/40 & 1/40 \\
Objaverse ($n=40$) & 1.98 & 0.43 & 0/40 & 0/40 \\
\midrule
All ($n=120$) & 3.11 & 0.97 & 12/120 & 1/120 \\
\bottomrule
\end{tabular}
\caption{\textbf{Mesh-only reference agreement under correct and swapped inputs.} Blinded
Astra judgments against Qwen3-VL-8B-Instruct captions, with the target reference and
instruction held fixed. Measured rows use the \lamdirect{} multitask checkpoint;
no \lamselect{} result in this table uses this caption-agreement rubric. The
newer \lamselect{} study uses blinded visual review (\cref{sec:vlxm10k}).
This is not a comparison between the two headline localization checkpoints.}
\label{tab:tdm-description}
\end{table}

\paragraph{Visual review.}
Before scoring, we fix $24$ meshes: eight evenly spaced sorted held-out IDs per
source. Blinded Astra reviews compare each description with three target renders,
which only the judge and reader see. The judge flags major contradictions in $15/24$
correct-mesh and $24/24$ swapped outputs; only $2/24$ correct-mesh outputs are
rated largely faithful. These are exploratory model judgments, not calibrated
accuracy. Of $16$ hidden controls, all eight negatives and two nominal positives
are flagged. The latter incorrectly call a U-shaped sofa L-shaped and remain
in the record as defective controls.

\paragraph{Pointing audits.}
On $32$ ABO cases with held-out object IDs in possibly seen scenes, the checkpoint
returns $26$ points, five refusals, and one malformed answer. Only $9/32$ hit the
target box, including four boundary hits: this is coarse localization, not
keypoint accuracy.

A real-image check selects $32$ YCB-V test and $32$ HB validation queries before
inference, spread across scenes and excluding HB's development scene. Targets
have at least $70\%$ visibility and visible boxes covering at least $2.5\%$ of
the image. With the same checkpoint, prompt, and decoder, all $64$ answers are
\texttt{<box>none</box>}. YCB-V supplies localization training only; these selected
targets do not define a representative benchmark.

\paragraph{Examples and scope.}
The compact galleries select successes and failures at $3{:}1$, not their audit
frequencies. Pointing hits lie strictly inside the target box and on the object;
description labels assess only the shown excerpts, not the full responses below.
The boot response has a minor shaft-shape error beyond its excerpt. Other
galleries retain fixed-print-set cases. Joint capability at the headline
localization accuracy remains untested.

\begin{figure}[p]
\centering
\includegraphics[width=0.9\textwidth]{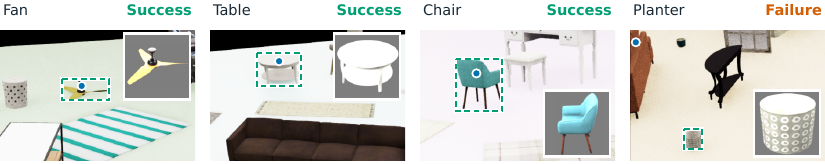}
\caption{\textbf{Pointing in synthetic ABO scenes.} The \lamdirect{} multitask checkpoint's blue points
hit three target boxes and miss one. Dashed green boxes mark the targets;
insets show the query meshes. These selected cases do not estimate accuracy.}
\label{fig:main-pointing}
\end{figure}

\begin{figure}[p]
\centering
\includegraphics[width=0.9\textwidth]{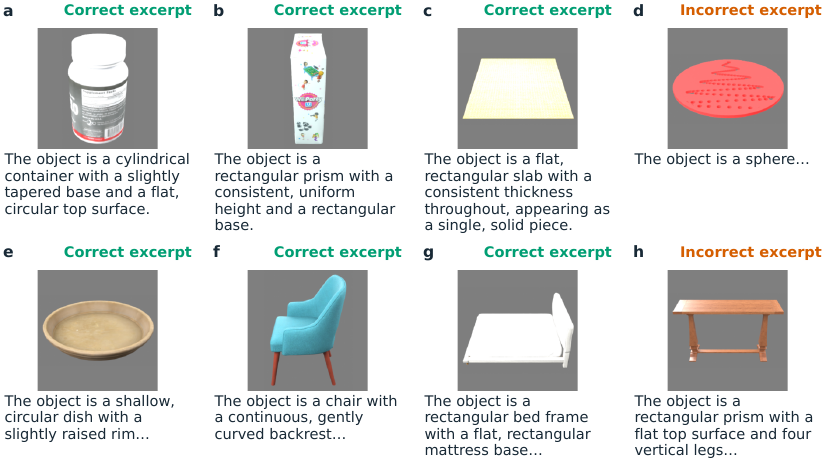}
\caption{\textbf{Mesh-only description.} Each row shows three correct excerpts and one
incorrect excerpt from the \lamdirect{} multitask checkpoint. Mesh renders are for the reader;
the model receives no image input. Full responses:
\cref{fig:main-description-full,fig:main-description-full-2}.}
\label{fig:main-description}
\end{figure}

\begin{figure}[p]
\centering
\includegraphics[width=\textwidth]{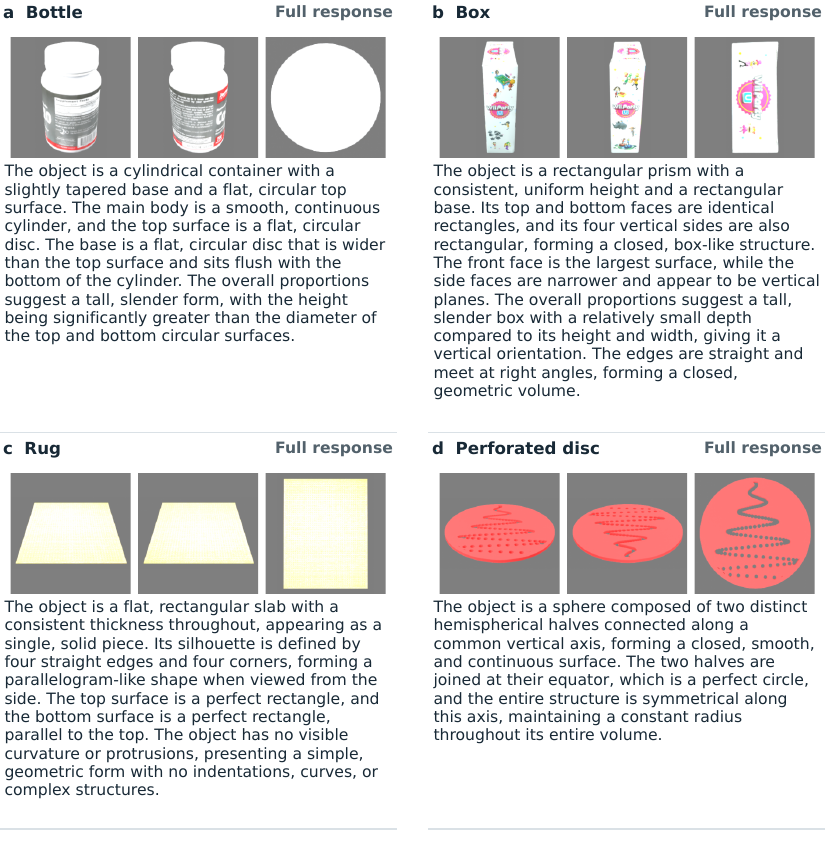}
\caption{\textbf{Full mesh descriptions (a--d).} The first description row of
\cref{fig:main-description}, with three views per mesh and complete literal responses.
Correctness labels in the compact gallery assess only the shown excerpts.}
\label{fig:main-description-full}
\end{figure}

\begin{figure}[p]
\centering
\includegraphics[width=\textwidth]{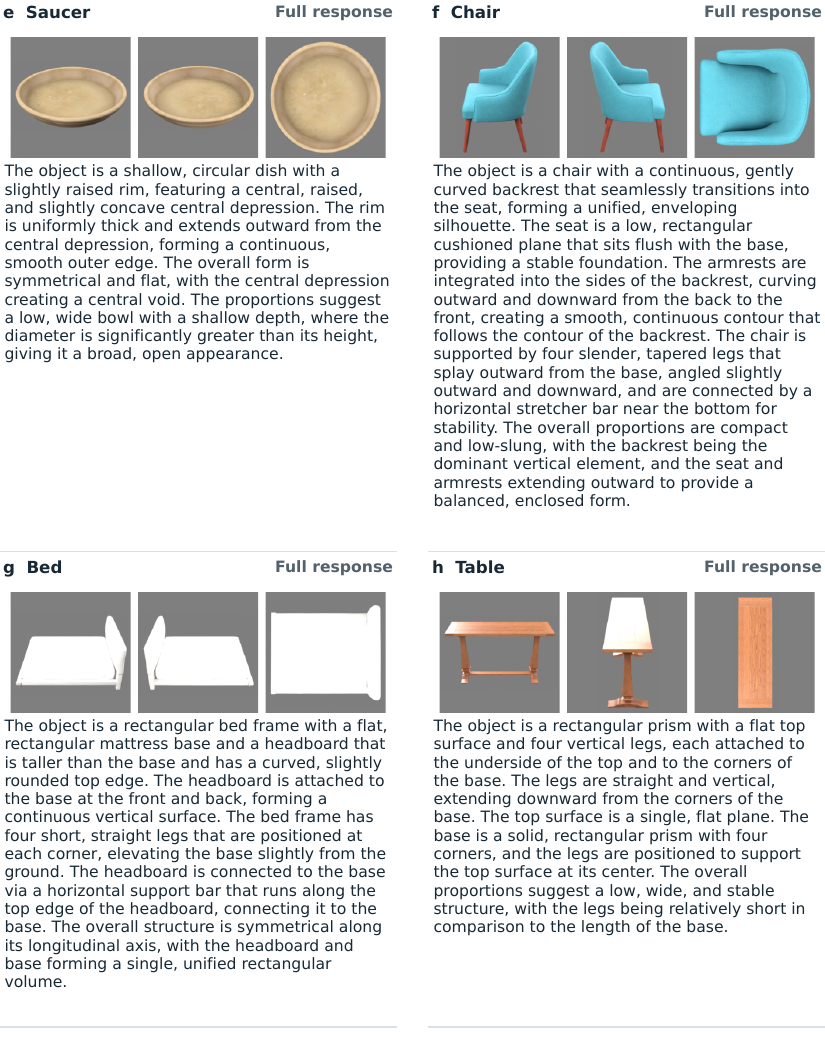}
\caption{\textbf{Full mesh descriptions (e--h).} The second description row of
\cref{fig:main-description}, in the same order. These selected examples do not replace the
fixed visual audit.}
\label{fig:main-description-full-2}
\end{figure}

\begin{figure}[p]
\centering
\includegraphics[width=\textwidth]{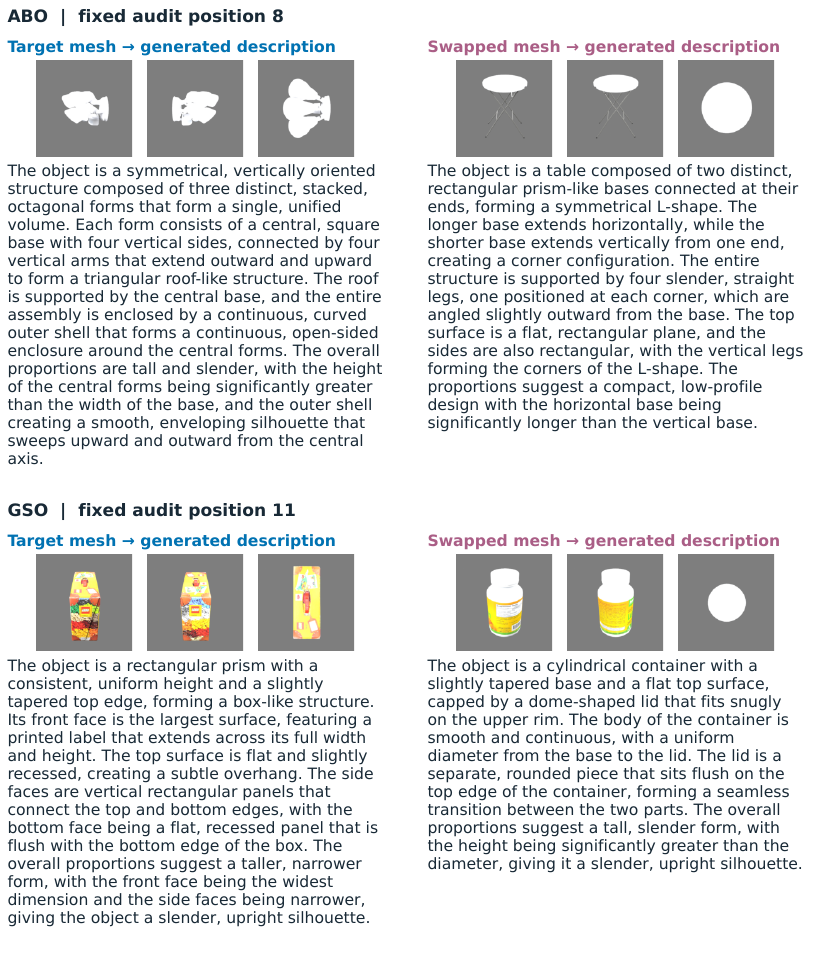}
\caption{\textbf{Mesh-only descriptions: ABO and GSO.} Multitask checkpoint, correct mesh (left)
and random swap (right), with full literal responses. Renders illustrate the inputs; the model
receives only mesh tokens. These are two cases from the fixed audit.}
\label{fig:capability-describe-b}
\end{figure}

\begin{figure}[p]
\centering
\includegraphics[width=\textwidth]{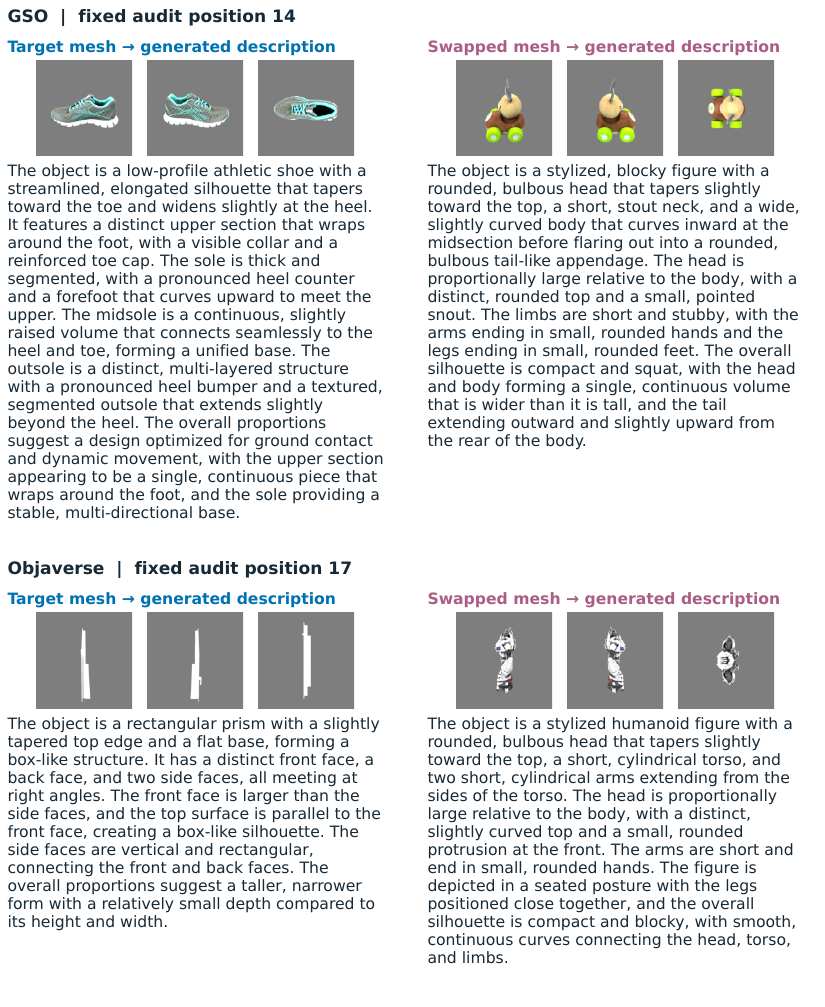}
\caption{\textbf{Mesh-only descriptions: GSO and Objaverse.} Multitask checkpoint, correct mesh
(left) and random swap (right), with full literal responses.
Renders are for the reader only. Together with \cref{fig:capability-describe-b}, these panels show
four cases from the fixed audit.}
\label{fig:capability-describe-c}
\end{figure}

\begin{figure}[p]
\centering
\includegraphics[width=\textwidth]{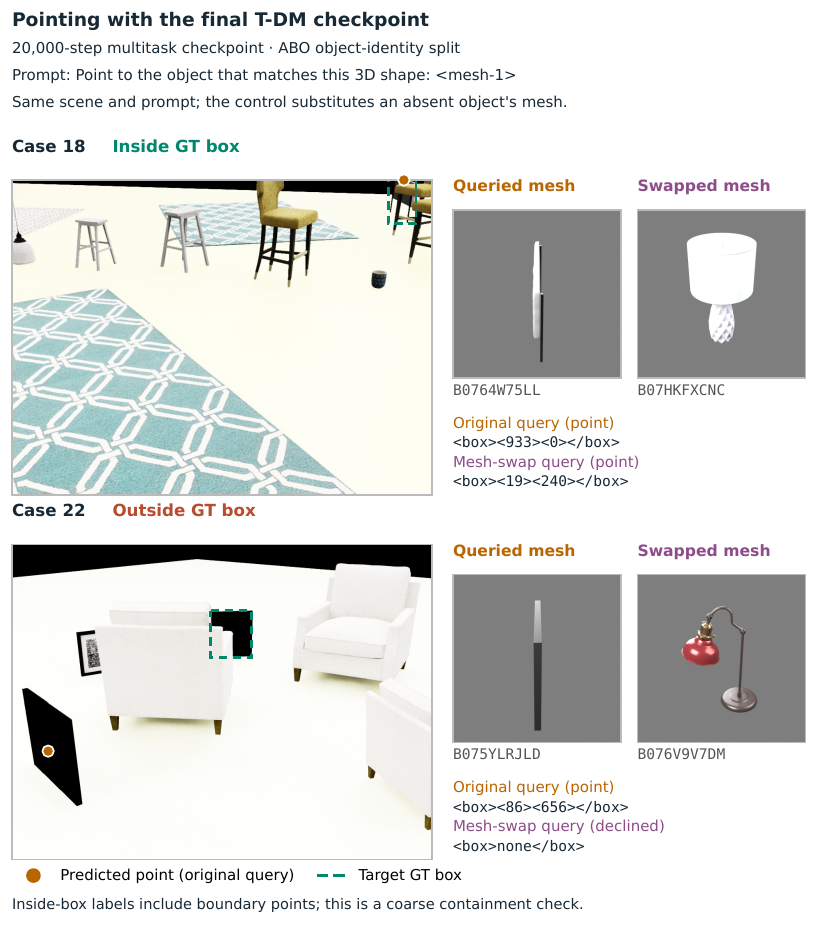}
\caption{\textbf{Pointing: a hit and a miss.} Multitask checkpoint, ABO cases $18$ and $22$ from
the fixed print set. Each scene and instruction is held fixed while the mesh is swapped. The
original-query point lies inside the target box in case $18$ and outside it in case $22$.
Literal outputs include the swapped-query responses. These selected cases do not estimate accuracy.}
\label{fig:capability-point-c}
\end{figure}

\FloatBarrier
\subsubsection{\lamselect{} M10k capability continuation}
\label{sec:vlxm10k}

M10k starts from R at update $31{,}434$ and runs for $10{,}000$ additional
updates with one seed. Its delivered examples follow an $80/15/5$
locate/describe/point mixture: $64{,}000$, $12{,}000$, and $4{,}000$ examples.
The run uses the same PPAT mesh representation. It is evaluated for capability
figures only; R remains the model in every localization result and table.

For mesh-only description, both R and M10k@10k receive the mesh tokens and an
instruction, with no image, regions, labels, caption, or reference render in the
model inputs. The panel contains $300$ held-out-caption identities. Ten blinded
assistant reviewers score correct and same-source, same-category swapped-mesh
outputs against three renders, with hidden positive and negative controls. A
grounded, non-generic response must contain at least one supported intrinsic
attribute, no major contradiction, no refusal, and no generic-only description.
Positive and negative controls reach $86.7\%$ and $5.7\%$ grounded. R scores
$52.7\%$ on correct meshes without description-task training; M10k@10k scores
$63.0\%$. Their paired difference is $10.3$ percentage points with a $95\%$
identity-grouped bootstrap interval of $[3.7,16.7]$. These are judge-assisted
visual-review results, not calibrated human accuracy.
Swapped-mesh grounded rates are $31.7\%$ for R and $44.3\%$ for M10k@10k;
correct-minus-swapped gaps are $21.0$ and $18.7$ percentage points, respectively.
These results do not establish stronger identity binding or a statistically
significant change in the gap.

For pointing, the fixed real-BOP panel contains $2{,}000$ present and $533$
absent queries. M10k@10k produces strict point syntax on $99.6\%$ of attempts,
hits the queried object's visible mask on $35.7\%$ of present attempts, and emits
a point on $26.6\%$ of absent attempts. Present hit rates are $60.6\%$ on YCB-V,
$17.8\%$ on LM-O, $29.6\%$ on HB validation, and $34.8\%$ on HOPE validation.
Coordinates decode with
$\operatorname{round}(x(W-1)/1000),\operatorname{round}(y(H-1)/1000)$.
Invalid, refused, and failed attempts stay in the denominators. Masks, crops,
and hit labels are evaluation-only and never enter the model.

\subsubsection{R native point-prompt transfer}
\label{sec:vlxpoint}

The $31{,}434$-step checkpoint receives the description probe's $32$ fixed
present-target image--mesh pairs, eight per suite, without proposals, labels,
ground-truth boxes, reference crops, or additional task training. One instruction
requests \texttt{<point>x,y</point>} with integer coordinates in $[0,1000]$, or
\texttt{<point>none</point>} if identification fails. Coordinates start at the
top-left, increasing rightward and downward, and map to image pixels as
$(xW/1000,yH/1000)$. Greedy generation permits $96$ new tokens with one attempt
per query, no prompt search or retries, and no conflicting grounding instruction.

\begin{figure}[p]
\centering
\includegraphics[width=\textwidth]{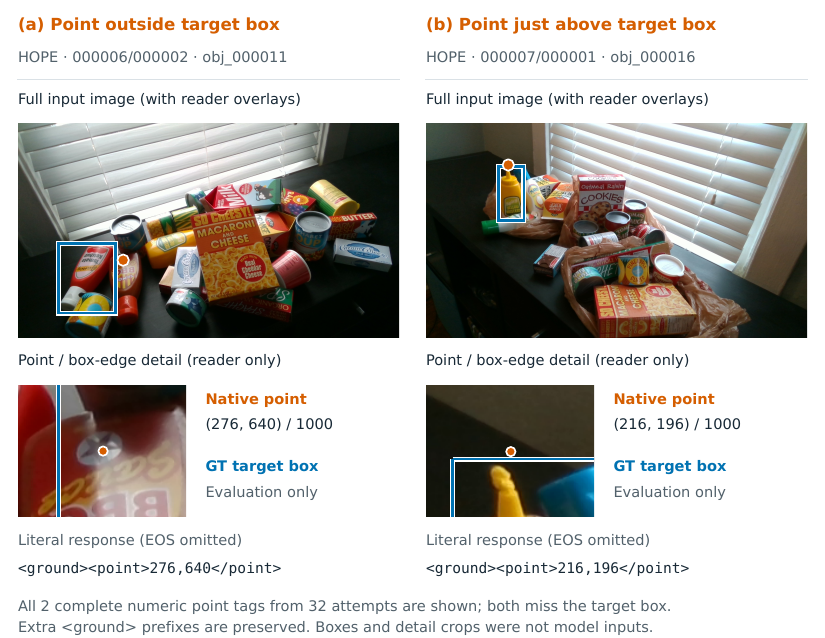}
\caption{\textbf{R native point-prompt failures.}
These are the only two answers containing complete point-coordinate tags among
$32$ fixed present-target queries to the localization checkpoint.
The displayed points come directly from the generated coordinates, scaled from
$[0,1000]$ to the original image dimensions. Both miss the target's visible box
and mask. Point markers, target-box overlays, and enlarged detail crops
are reader aids; the model receives only the unannotated image, mesh features,
and instruction. Responses are literal, including the incomplete grounding
wrapper; only the end-of-sequence marker is omitted. No answer meets the exact
requested format. This one-prompt transfer test does not establish general
pointing performance.}
\label{fig:vlx-point-probe}
\end{figure}

There are $0/32$ exact-format answers. Two contain complete point tags preceded
by an unclosed \texttt{<ground>} wrapper; the other $30$ have no complete point
tag. Both emitted points miss the manifest boxes and original HOPE visible/full
boxes and masks (\cref{fig:vlx-point-probe}); inspection neither repairs wrappers
nor extracts points from free-form numbers. This single-prompt, present-target
probe provides no successful native pointing example, does not test absence,
and does not establish a general \lamselect{} pointing result or isolate the cause
of the retained grounding syntax.

\subsubsection{R image-plus-mesh description}
\label{sec:vlxdescription}

The main \lamselect{} + PPAT checkpoint at $31{,}434$ steps receives an image,
$65$ mesh tokens, and an instruction to describe the queried object's visible
appearance or abstain. There is no additional mesh-description supervision;
labels, ground-truth boxes, reference crops, and proposal candidates are excluded.
Greedy decoding permits $96$ new tokens. This is image-conditioned, not mesh-only
description.

The cohort has eight original present queries per suite (HOPE, LM-O, HB, YCB-V),
same-image substitutions with another present mesh, and absent-image controls
(\cref{tab:vlx-description}). All $96$ answers are nonempty and EOS-terminated,
but $95/96$ retain grounding markup. All $32$ present-mesh substitutions change
the answer, establishing sensitivity rather than successful identification.

\begin{table}[htbp]
\centering
\small
\setlength{\tabcolsep}{4pt}
\begin{tabular}{lrcrr}
\toprule
Condition & $n$ & \makecell{Identity\\S / U / C} &
\makecell{Refusal-like\\text} & \makecell{Appropriate\\abstention} \\
\midrule
Original present mesh & 32 & 11 / 18 / 3 & 0 & \na \\
Other present mesh & 32 & 7 / 17 / 4 & 4 & \na \\
Absent image, same suite & 25 & \na & 16 & 13 \\
Absent image, cross suite & 7 & \na & 6 & 5 \\
\bottomrule
\end{tabular}
\caption{\textbf{Exploratory \lamselect{} image-plus-mesh description controls.}
Counts use all $n$ attempts in each condition. Assistant identity judgments are
supported (S), uncertain (U), or contradicted (C); the four remaining present-mesh
substitutions are refusal-like outputs with identity rated not applicable.
Appropriate abstention counts only affirmative assistant judgments, retaining
negative, uncertain, and not-applicable cases in $n$.
Refusal-like text includes bare \texttt{no}/\texttt{None}, but excludes
\texttt{<objects>None</objects>}: no proposals were supplied, so this field is
not evidence of abstention. \na\ denotes a measure not reported for that condition.
These are review counts, not calibrated human accuracy.}
\label{tab:vlx-description}
\end{table}

\begin{figure}[p]
\centering
\includegraphics[width=\textwidth]{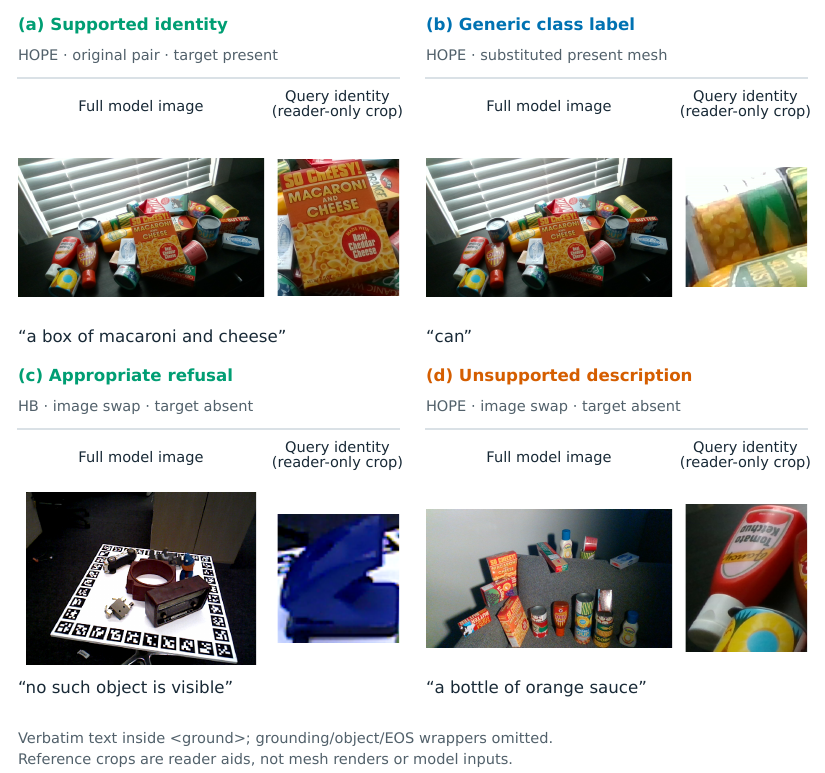}
\caption{\textbf{R descriptions: supported, generic, and unsupported responses.}
Each panel shows the complete input image and a positive-reference crop of the
queried identity for the reader only; the crop is neither a mesh render nor a
model input. Panels (a,b) hold the image and instruction fixed while substituting
another present mesh: a supported label becomes a generic answer, whose identity
is uncertain rather than demonstrably wrong. Panels (c,d) show an appropriate
refusal and an unsupported description under same-suite absent-image controls.
Text is the verbatim content of \texttt{<ground>}, with grounding/object wrappers
and the end-of-sequence marker omitted. Examples were selected after review by
the lowest review ID within these outcome strata, pairing (a) with its mesh
substitution; they illustrate behavior, not its frequency.}
\label{fig:vlx-descriptions}
\end{figure}

Each item has one assistant review, with partial blinding and no inter-rater
calibration. Reviewers see the input image and reviewer-only target-reference
crops and visible-instance masks. Identity support does not guarantee correct
attributes. Presence follows dataset-qualified raw annotations, including
invisible instances; cross-suite controls change image domain and lack an
exhaustive physical-identity crosswalk. These selected, dependent controls are
not a representative score, and no matched \lamdirect{} or Uni3D run uses this
protocol. \Cref{fig:vlx-descriptions} shows post-hoc selected examples.

\end{document}